\documentclass[conference]{IEEEtran}
\usepackage{amsmath}
\usepackage{marvosym} 
\usepackage{url}
\usepackage{bbold}
\usepackage{graphicx}  
\usepackage{float}     
\usepackage{makecell}
\usepackage{booktabs} 
\usepackage{multirow}
\usepackage[table]{xcolor}
\usepackage[caption=false,font=small,labelfont=rm,textfont=rm]{subfig}
\usepackage{enumitem}
\usepackage{tcolorbox}
\usepackage{pifont}
\usepackage{listings}
\usepackage{booktabs}
\usepackage{multirow}
\usepackage{cite}
\usepackage{ulem}
\usepackage{pifont}

\newenvironment{mybox}{\begin{tcolorbox}}{\end{tcolorbox}}

\newcommand{\sysname}{{SATA}}

\ifCLASSINFOpdf
\else
\fi
\begin{document}
%



\title{More Than Meets the Eye: A Semantics-Aware Traffic Augmentation Framework for Generalizable Website Fingerprinting}




\author{\IEEEauthorblockN{
Youquan Xian\IEEEauthorrefmark{1}\IEEEauthorrefmark{4}\IEEEauthorrefmark{6}\thanks{\IEEEauthorrefmark{6} Equal contribution.}, 
Xueying Zeng\IEEEauthorrefmark{2}\IEEEauthorrefmark{6}, 
Lingjia Meng\IEEEauthorrefmark{4}, 
Lei Cui\IEEEauthorrefmark{4}\textsuperscript{\Letter}, 
Runhan Song\IEEEauthorrefmark{4}\IEEEauthorrefmark{5}, 
Wei Wang\IEEEauthorrefmark{4}, \\
Zhengquan Ding\IEEEauthorrefmark{4},  
Peng Liu\IEEEauthorrefmark{3}, 
Zhiyu Hao\IEEEauthorrefmark{4}\textsuperscript{\Letter} \thanks{\textsuperscript{\Letter}Corresponding author.}}

\IEEEauthorblockA{\IEEEauthorrefmark{1}School of Cyberspace Security, Beijing University of Posts and Telecommunications, Beijing, China}
\IEEEauthorblockA{\IEEEauthorrefmark{2}School of Computer Science and Engineering, Beihang University, Beijing, China}
\IEEEauthorblockA{\IEEEauthorrefmark{5}Faculty of Computing, Harbin Institute of Technology, Harbin, China}
\IEEEauthorblockA{\IEEEauthorrefmark{3}School of Computer Science and Engineering, Guangxi Normal University, Guilin, China}
\IEEEauthorblockA{\IEEEauthorrefmark{4}Zhongguancun Laboratory, Beijing, China}
}



\IEEEoverridecommandlockouts
\makeatletter\def\@IEEEpubidpullup{6.5\baselineskip}\makeatother
\IEEEpubid{\parbox{\columnwidth}{
		Network and Distributed System Security (NDSS) Symposium 2026\\
		23 - 27 February 2026 , San Diego, CA, USA\\
		ISBN 979-8-9919276-8-0\\  
		https://dx.doi.org/10.14722/ndss.2026.[23$|$24]xxxx\\
		www.ndss-symposium.org
}
\hspace{\columnsep}\makebox[\columnwidth]{}}

\maketitle

\begin{abstract}

Deep learning-based website fingerprinting has emerged as an effective technique for inferring the websites users visit. Although existing methods achieve strong performance on closed-world datasets, they often fail to generalize to real-world environments, especially under geographic and temporal shifts. This limitation fundamentally stems from the coupled effects of two key challenges: application-layer resource composition variability and observable feature instability induced by cross-layer encapsulation. Intertwined, these factors induce systematic shifts between underlying application semantics and observable traffic features. To address the above challenges, we propose {\sysname}, a semantics-aware traffic augmentation framework. Specifically, {\sysname} first performs application-layer semantic augmentation based on protocol rules, expanding the resource composition patterns within each flow and frame sequence patterns under protocol constraints. Based on these augmented frame sequences, we further introduce a cross-layer feature alignment mechanism via knowledge distillation. It aligns frame sequence with packet-length sequence features, enabling cross-layer feature alignment between enhanced semantics and observable sequences.
Extensive experiments show that {\sysname} successfully generates traffic patterns that are absent from the training set but genuinely exist in the test set, and significantly improves the performance of mainstream models across diverse and complex scenarios. In particular, in open-world settings, {\sysname} improves ACC by 90.81\% and AUROC by 48.37\%. The source code of the prototype system is available at \url{https://anonymous.4open.science/r/SATA-B6C2/}.

\end{abstract}

%


%
\IEEEpeerreviewmaketitle


\section{Introduction}
In recent years, end-to-end encrypted protocols, exemplified by TLS 1.3 \cite{lee2021tls}, along with privacy-enhancing technologies such as encrypted DNS \cite{rfc8484,rfc8310,rfc9250} and Encrypted Client Hello \cite{rfc9849}, have fundamentally reshaped the privacy landscape of cyberspace. The widespread deployment of these mechanisms has rendered traditional traffic analysis methods that rely on deep packet inspection and plaintext metadata increasingly ineffective \cite{cdn,shen2022machine,papadogiannaki2021survey,wickramasinghe2025sok,zhao2025sweet}. However, protocol-level encryption does not eliminate the risk of privacy leakage. During communication, side channels such as packet length and packet direction can still expose observable features \cite{sharma2025survey,feng2025unmasking}. With their strong capability for non-linear representation learning, advanced website fingerprinting (WF) models can bypass payload encryption and infer high-level application behaviors solely from these observable signals, thereby posing a new threat to network privacy \cite{bahramali2023realistic,hayes2016k,sirinam2018deep,van2020flowprint,qu2023input}.



\begin{figure}[t]
\centering
\includegraphics[width=\linewidth]{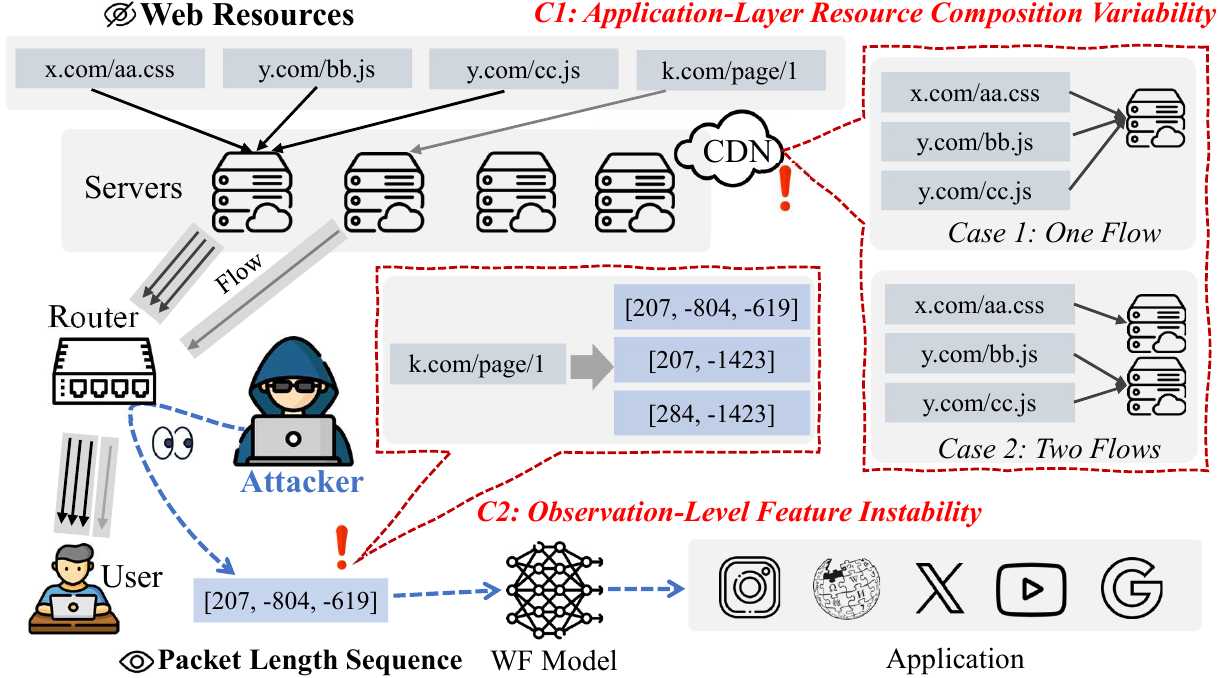}
\caption{Overview of a WF attack scenario and its two major generalization challenges. Blue dashed lines denote the attacker’s workflow, and red dashed regions indicate the challenges faced by the attacker.}
\label{fig:intro}
\end{figure}

Although existing WF models have achieved promising performance on closed-world datasets, variations in resource scheduling and multilayer protocol encapsulation mechanisms in real networks pose severe generalization challenges in the real world.
To improve generalization, existing studies have primarily relied on the data augmentation paradigm, addressing the problem from two perspectives: data distribution fitting and network simulation. On the one hand, methods based on deep generative models such as generative adversarial networks (GANs) and large language models (LLMs) are mostly limited to generating intermediate representations or non-functional fields \cite{alomar2025detection,jiang2024netdiffusion,zhu2023iletc,sun2025advtg,hajaj2024art}. Such purely data-driven paradigms are highly dependent on the training set's original distribution and struggle to extend beyond the existing probability space to generate plausible out-of-domain (OOD) data \cite{gan_bug,dg_bug}.
On the other hand, existing simulation-based augmentation strategies, such as those simulating perturbations in round-trip time (RTT) and maximum transmission unit (MTU), remain confined to TCP/IP stack–level effects \cite{zhao2025nuwa,zion2025enhancing,xie2023rosetta,bahramali2023realistic,horowicz2022few}. More importantly, both paradigms overlook how application-layer resource scheduling and cross-layer encapsulation jointly shape traffic features. As a result, the mapping from application semantics, namely the actual conveyed resources, to observed traffic features exhibits a systematic shift that is difficult to correct.

We observe that the systematic shift of observed traffic features fundamentally arises from the coupled effects of two challenges during traffic generation, as illustrated in Fig.~\ref{fig:intro}.
First, at the application layer, the interplay of dynamic DNS \cite{wang2018evolution,schomp2020akamai} and HTTP/2 multiplexing \cite{rfc7540} continuously alters the composition of resources within a flow, resulting in resource composition variability (C1).
Second, during cross-layer encapsulation, application resources undergo structural perturbations induced by state-dependent header compression \cite{rfc7541}, fragmentation, and dynamic scheduling mechanisms \cite{rfc7540,rfc9293,rfc5246}, resulting in unstable observed features, such as packet-length sequences (C2). The systematic feature shift induced by these challenges significantly degrades the generalization capability of WF models in real-world environments.


To address the aforementioned challenges, this paper provides an analysis of the multilayer factors underlying the systematic shift of observed traffic features. On this basis, we propose a semantics-aware traffic augmentation framework, {\sysname}. Specifically, {\sysname} first expands the resource composition patterns within each flow and the application-layer frame sequence patterns under protocol constraints through a resource recomposition module and a frame sequence augmentation module. It then introduces a cross-layer feature alignment mechanism that uses frame sequence representations to guide the learning of packet-length sequences. This design mitigates the systematic shift of observed features, improving the generalization and robustness of existing WF models in the real world. 
To the best of our knowledge, this is the first work to leverage application-layer semantics for traffic augmentation.

The main contributions are summarized as follows:
\begin{itemize}
    \item We analyze the multilayer factors underlying the systematic shift of observed traffic features, revealing how application-layer behaviors and cross-layer protocol mechanisms jointly reshape traffic features. It provides a theoretical foundation for understanding the mechanisms driving such feature shifts.
    \item We propose a semantic-aware traffic augmentation framework, {\sysname}, which simulates realistic network protocol processes to enrich application-layer frame sequences, while also constructing a cross-layer feature alignment mechanism to align observed features with semantic features.
    \item Extensive experimental results validate the effectiveness of {\sysname}. It improves ACC and AUROC by 90.81\% and 48.37\% in open-world settings, respectively, and increases pattern coverage by 9.93\%, generating frame sequence patterns unobserved in the training set but present in the test set.
\end{itemize}

\section{Related Work}
\label{bg}

\subsection{Advances in Website Fingerprinting}

With the widespread adoption of end-to-end encryption, the focus of WF research has transitioned from labor-intensive, pioneering manual feature engineering \cite{hayes2016k,taylor2016appscanner,van2020flowprint,shen2019encrypted} to deep learning-based automatic representation learning. The selection of model input features has been extensively investigated in previous works. Although some recent studies directly learn from raw payload bytes in an end-to-end manner \cite{lin2022bert,peng2024ptu,wang2024netmamba,chen2025miett,zhou2025trafficformer,peng2025bottom,zhao2023yet,liu2024atvitsc,zhang2023tfe,li2025sat,han2024gnn,zhang2025revolutionizing}, the latest research demonstrates that the pseudo-randomness of encrypted payloads can easily induce such models to rely on dataset-specific biases through shortcut learning~\cite{zhao2025sweet}. In contrast, packet length sequences and associated side-channel features, such as timing and direction, circumvent payload-level obfuscation and have been shown to provide more fundamental and reliable representations. They have therefore become the central focus of current WF research~\cite{wickramasinghe2025sok}.

For sequence features such as packet length, researchers have developed a wide range of deep learning models and architectures at different granularities. From the perspective of granularity, flow-level methods leverage models such as CNN and LSTM to perform lightweight modeling of single-flow features \cite{liu2019fs,piet2023ggfast,chen2020length,chen2022length,wu2022online,cai2022metc,bertps}. 
Trace-level methods, by contrast, incorporate more complex architectures to integrate global session context \cite{sirinam2018deep,li2021rltree,chen2023classify,fu2024detecting,qu2023input,xian2025udfs}. Benefiting from these powerful deep neural architectures and refined feature engineering, existing WF models generally achieve remarkable classification accuracy on public closed-world datasets, with performance nearing saturation. 
However, such performance advantages obtained in controlled environments frequently collapse in the wild~\cite{dong2025deep,bahramali2023realistic,xie2023rosetta}.

\subsection{Data Augmentation}
Recent studies have widely adopted data augmentation strategies to expand the training space, thereby addressing the generalization bottlenecks in existing models. The mainstream strategies can be broadly categorized into two categories: data distribution fitting and network simulation.

The first category relies on data-driven distribution fitting and extensively employs deep generative models, such as generative adversarial networks (GANs), diffusion models, and large language models (LLMs), to synthesize and augment traffic representations. 
Specifically, to mitigate class imbalance, ILETC \cite{zhu2023iletc}, CS-BiGAN \cite{alomar2025detection}, and NetDiffusion \cite{jiang2024netdiffusion} generate samples with high statistical similarity by directly learning traffic sequences or implicitly fine-tuning image-like representations to accurately capture the latent distribution of traffic. Hajaj et al. \cite{hajaj2024art} employed LSTMs to perform temporal extrapolation on traffic feature images. AdvTG \cite{sun2025advtg}, in contrast, leverages a fine-tuned LLM to selectively mutate non-functional fields in payloads, achieving semantic-level adversarial augmentation. However, distribution fitting methods often overlook the generation logic of traffic features, making it difficult for existing generative models to synthesize an interpretable, valid OOD packet length sequences~\cite{gan_bug,dg_bug}.

The second category augments traffic representations by manually manipulating network states, such as MTU and RTT, or by simulating packet loss and packet reordering, to force models to learn invariant features under diverse transmission conditions. Specifically, Rosetta \cite{xie2023rosetta} and Zion et al. \cite{zion2025enhancing} dynamically adjust network transmission parameters to generate synthetic samples characterized by sequence shifts and size variations. Horowicz et al. \cite{horowicz2022few} further extend such delay and packet-loss perturbations to customized augmentation of two-dimensional traffic images.
At the granularity of structural operations on packet sequences, NetAugment \cite{bahramali2023realistic} and Zion et al. \cite{zion2025enhancing} simulate feature shifts caused by bandwidth fluctuations through fine-grained modification of burst sequences and averaging of cross-flow features, respectively. 
Meanwhile, Nuwa \cite{zhao2025nuwa} further introduces masking strategies for packet loss and reordering, leveraging self-supervised learning to reconstruct corrupted complete features dynamically. However, these methods mainly operate on TCP/IP-stack-level observations, without touching the application-layer semantics that generate traffic.


Therefore, it is imperative to develop a traffic augmentation mechanism that bridges application semantics and observable features to enhance model generalization in the real world.

\section{Challenges}
\label{sec:motivation}




Expanding on the challenges previously outlined,  we examine the multilayer factors driving the systematic shift of traffic features from two perspectives: \textbf{(C1) Application-Layer Resource Composition Variability}, and \textbf{(C2) Observation-Level Feature Instability} induced by similar application semantics.


\subsection{Application-Layer Resource Composition Variability}

\begin{figure}[h]
    \centering
    \includegraphics[width=\linewidth]{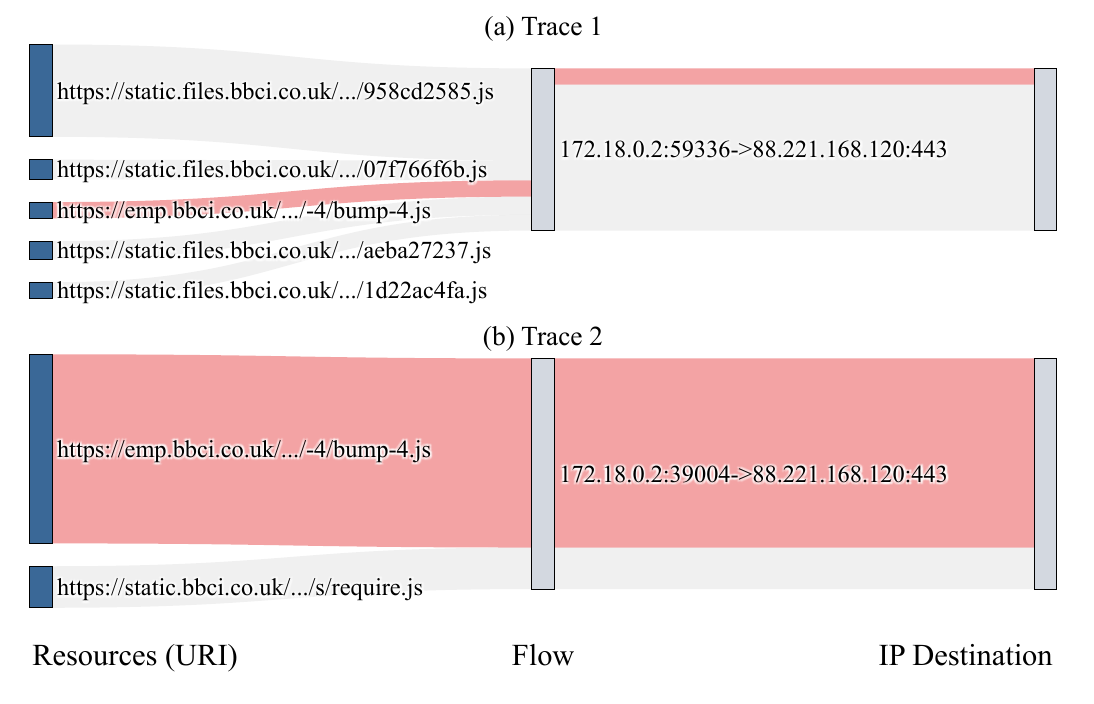}
    \caption{Illustrative example of cross-domain resource aggregation caused by dynamic DNS and HTTP/2 connection coalescing.}
    \label{fig:cdn}
\end{figure}

\begin{figure}[h]
    \centering
    \includegraphics[width=\linewidth]{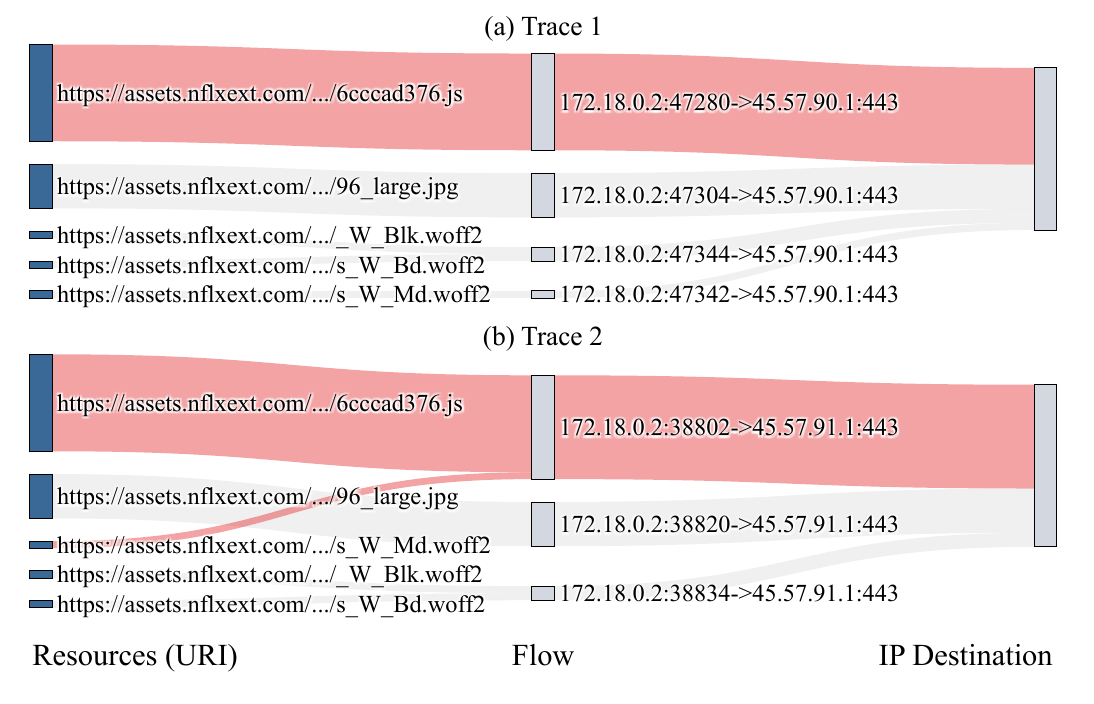}
    \caption{Illustrative example of flow-level resource distribution variation induced by HTTP/2 connection reuse.}
    \label{fig:h2_cconcurent}
\end{figure}


Regarding resource composition variation, two representative scenarios are observed, as illustrated in Fig.~\ref{fig:cdn} and Fig.~\ref{fig:h2_cconcurent}, namely changes in domain combination within a flow and the absence of multiplexing for same-domain resources within a single flow.

The first scenario arises from the interplay of dynamic DNS, shared infrastructure, and HTTP/2 connection coalescing. In practice, diverse domains may be resolved to the same service endpoint via CDNs~\cite{cdn} or reverse proxies \cite{reese2008nginx}. When these domains satisfy connection reuse criteria, such as TLS certificate compatibility, the client may leverage a single TCP connection to transmit logically independent cross-domain requests. Consequently, resources from multiple domains are consolidated into one flow, altering the original domain composition and traffic distribution patterns.
As illustrated in Fig.~\ref{fig:cdn}, across different visits, resources of domain \texttt{emp.bbci.co.uk} may co-occur within the same flow as \texttt{static.files.bbci.co.uk} or \texttt{static.bbci.co.uk} due to HTTP/2 connection coalescing, leading to variations in resource composition within the flow.

The second scenario primarily arises from the inherent non-determinism of HTTP/2 connection reuse strategies in practical implementations. Although HTTP/2 recommends reusing existing connections to enhance transmission efficiency, RFC 9113 explicitly states that clients are not mandated to enforce reuse~\cite{rfc9113}. 
The client may dynamically establish new TCP connections or shift scheduling across multiple connections under various triggers, such as existing connection loads approaching their thresholds, active termination by either party, or intervention by browser security isolation policies.
Consequently, resource requests under the same domain may be distributed across multiple parallel flows, manifesting a multi-flow transmission pattern. As illustrated in Fig.~\ref{fig:h2_cconcurent}, the red-marked resource from \texttt{assets.nflxext.com} occupies an exclusive flow in one visit, while in another it shares a flow with other resources, leading to significantly different transmission structures (Appendix~\ref{appendx_c1}).
This volatility decouples flows from semantics, rendering models overfitted to static training distributions fragile against the real world.

\begin{figure}[h]
    \centering
    \includegraphics[width=\linewidth]{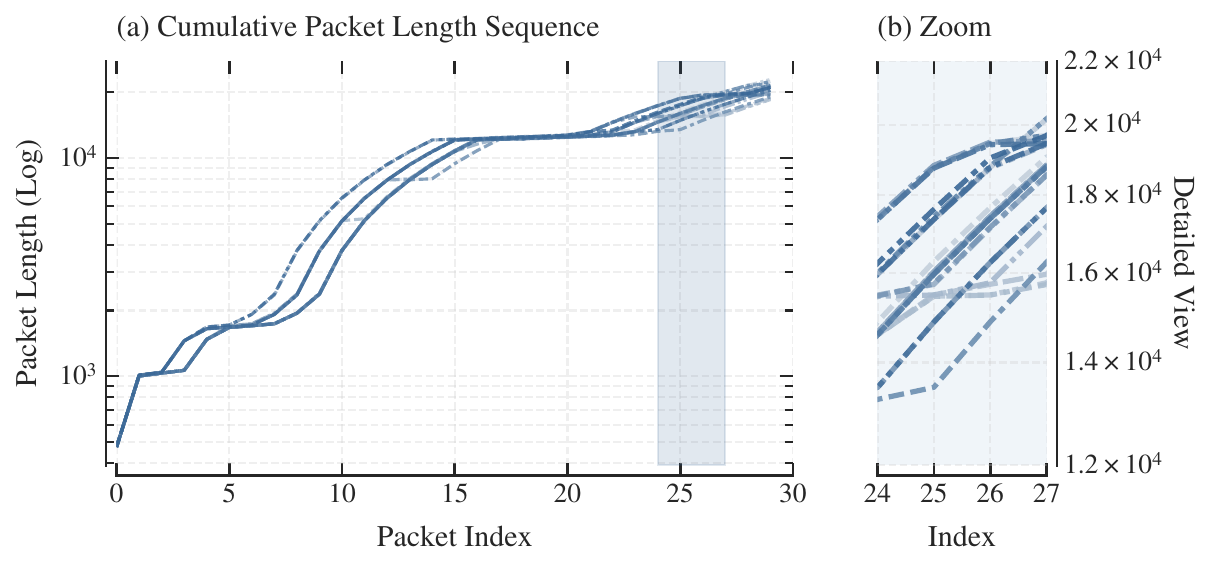}
    \caption{Illustrative example of packet length sequence instability caused by HTTP/2 scheduling and cross-layer encapsulation.}
    \label{fig:flow_pkl}
\end{figure}

\subsection{Observation-Level Feature Instability}

Beyond variations in resource composition, packet length sequences of the same resource can exhibit significant variations, as illustrated in Fig.~\ref{fig:flow_pkl}. In the mapping from application-layer resources to HTTP/2 frames, such uncertainty primarily originates from state dependencies and scheduling coupling within the protocol. On the one hand, \textit{HPACK}, the header compression mechanism used in HTTP/2, leverages a stateful dynamic table to reduce redundant header transmission. Its stateful nature leads \texttt{HEADER} frames to alternate between indexed header field representation and literal encoding, resulting in variable frame sizes \cite{rfc7541}. On the other hand, the concurrent requests of partial resources under multiplexing drives the protocol stack to transmit \texttt{HEADERS} frames in compacted bursts, disrupting the expected sequential structure of \texttt{HEADERS} and \texttt{DATA} frames in observed traces (Appendix~\ref{appendx_c3}).




Furthermore, during cross-layer encapsulation, buffering and asynchronous scheduling across the HTTP, TLS, and TCP layers undermine the stability of data unit boundaries as they propagate through the protocol stack. Dynamic buffer write patterns, evolving window states, and transmission scheduling jointly drive continuous fragmentation of data units, resulting in non-linear segmentation in the top-down mapping of application-layer semantics (Appendix~\ref{appendx_c2}). Finally, at the transport layer, the mapping from TLS records to TCP segments is further modulated by factors such as maximum segment size (MSS) \cite{rfc879} negotiation, exacerbating this cross-layer structural perturbation.
It disrupts learned temporal dependencies, constraining the generalization capability of conventional WF models in the real world.


\begin{figure*}[t!]
    \centering
    \includegraphics[width=\linewidth]{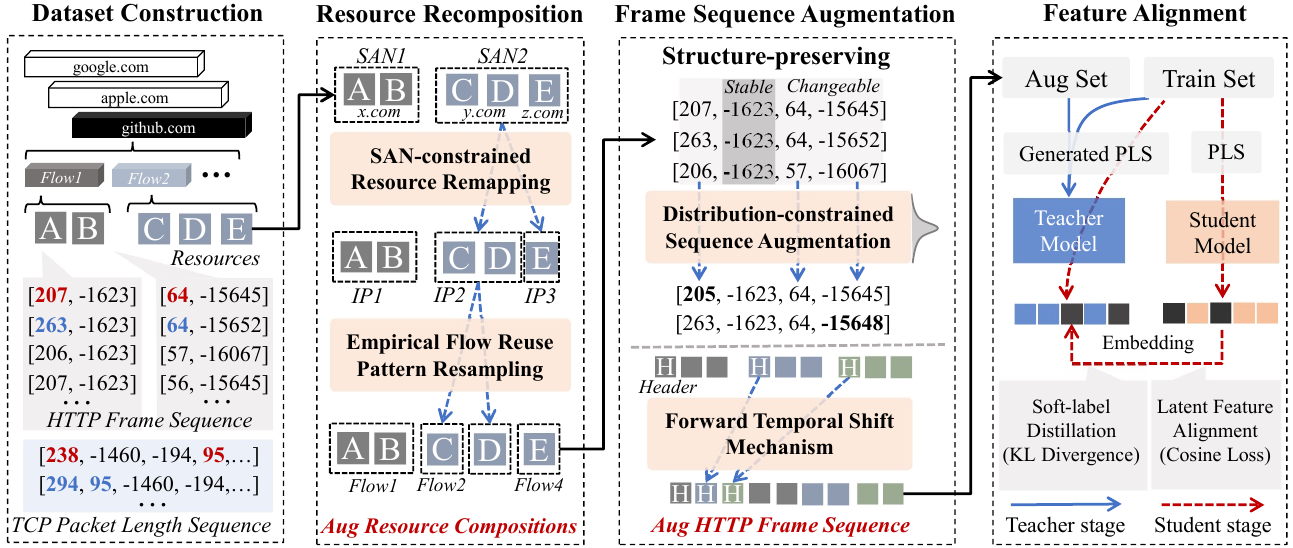}
    \caption{The workflow of {\sysname} is as follows: (1) Dataset Construction module establishes precise alignment between plaintext resources and encrypted traffic, providing the data foundation. (2) Resource Recomposition module operates on this dataset to simulate dynamic DNS and HTTP/2 multiplexing, generating flows with diverse resource composition patterns. (3) Frame Sequence Augmentation module perturbs the transmission volume and positional distribution of frame sequences within the recombined flows, and models header coalescing behavior, expanding frame sequence patterns. (4) The Feature Alignment module employs knowledge distillation to transfer semantically aligned representations from augmented sequences into the WF model.}
    \label{fig:overview}
\end{figure*}

\section{System Design}
\label{methodology}

\subsection{Threat Model}
We consider a website fingerprinting scenario in real-world HTTP/2 traffic, where deep learning models are deployed in operational networks to perform classification based solely on observable packet-length sequences extracted from TCP flows. However, modern web infrastructure and the HTTP/2 protocol jointly lead to significant variability in traffic features. These factors collectively induce significant distribution shifts between training and deployment environments, impeding the generalization of existing models in realistic settings.

{\sysname} aims to improve the classification robustness of existing deep learning models when handling real-world HTTP/2 traffic. It does not require prior knowledge of the network environment. In practice, mechanisms such as dynamic scheduling and protocol stack encapsulation on both the client and server sides are highly time-varying and largely unobservable, making them extremely difficult to capture accurately in real time. Moreover, {\sysname} maintains compatibility with existing models without requiring architectural modifications. It only necessitates a knowledge distillation framework, where the original WF model is instantiated as both teacher and student, and trained in a two-stage scheme across distinct tasks.

\subsection{Overview}

This paper presents {\sysname}, a semantics-aware traffic augmentation framework, as illustrated in Fig.~\ref{fig:overview}. {\sysname} is designed to mitigate the systematic shift stemming from both application-layer resource composition variability (C1) and observation-level feature instability (C2), thereby enhancing the robustness of mainstream deep learning models on real-world HTTP/2 traffic.

{\sysname} establishes a systematic pipeline spanning dataset construction, traffic augmentation, and feature alignment. 
Specifically, it first formalizes a semantic correspondence between plaintext resources and encrypted traffic. Subsequently, by incorporating resource recomposition and frame-sequence augmentation tailored to mechanisms like dynamic DNS and HTTP/2 multiplexing, the framework diversifies flow-level application semantics to encompass the varied traffic patterns encountered in operational networks. 
Finally, a knowledge distillation-based feature alignment mechanism transfers semantic knowledge from frame sequences to the observations packet-length representation, enabling the model to inherit the enriched semantics while mitigating systematic shift. {\sysname} maintains full compatibility with existing architectures; it involves instantiating the original WF model as both a teacher and a student within a two-stage knowledge distillation paradigm. In the following subsections, we introduce the four core modules of {\sysname} in detail.

\subsection{Dataset Construction}
\label{subsec:data_prep}

To address the lack of fine-grained alignment between plaintext resources and encrypted observations in existing encrypted traffic datasets, this paper leverages \texttt{Tshark} \footnote{\url{https://tshark.dev/}} and TLS session keys to perform cross-layer parsing of raw traffic, thereby establishing correspondences among resources, HTTP/2 frame sequences, and packet length sequences. Specifically, two types of mappings are constructed. First, TLS traffic is decrypted to extract the HTTP/2 frame sequence corresponding to each resource, consisting of the sizes of \texttt{HEADER} and \texttt{DATA} frames, establishing a mapping between resources and frame sequences. Second, at the flow level, resource compositions and their corresponding TCP packet length sequences are extracted to establish a mapping between resource compositions and observable traffic representations. Through this process, a cross-layer aligned dataset linking resources, frame sequences, and packet length sequences is constructed, providing the data foundation for subsequent resource recomposition, frame sequence augmentation, and cross-layer feature alignment.

\subsection{Resource Recomposition}
\label{subsec:flow_restructure}

\definecolor{codegreen}{rgb}{0,0.6,0}
\definecolor{codegray}{rgb}{0.5,0.5,0.5}
\definecolor{codepurple}{rgb}{0.58,0,0.82}
\definecolor{backcolour}{rgb}{0.95,0.95,0.92}

\lstdefinestyle{mystyle}{
  backgroundcolor=\color{backcolour}, 
  commentstyle=\color{codegreen},
  keywordstyle=\color{magenta},
  numberstyle=\tiny\color{codegray},
  stringstyle=\color{codepurple},
  basicstyle=\ttfamily\scriptsize,
  breakatwhitespace=false,         
  breaklines=true,                 
  captionpos=b,                    
  keepspaces=true,                 
  numbers=left,                    
  numbersep=5pt,                  
  showspaces=false,                
  showstringspaces=false,
  showtabs=false,                  
  tabsize=2
}

\lstset{style=mystyle}

\begin{lstlisting}[language=C++, caption=HTTP/2 reuse logic via certificate SAN validation (Chromium)., label=list1]
bool VerifyDomainAuthentication(domain) {
    if (session_is_draining)
        return false;
    if (!GetSSLInfo())
        return true;  // non-TLS session
    return CanPool(domain);
}
bool CanPool(new_hostname) {
    if (IsCertStatusError())
        return false;
    if (!cert->VerifyNameMatch(new_hostname))
        return false;  // SAN validation (core constraint)
    if (IsPKPViolated())
        return false;
    return true;
}
\end{lstlisting}

As shown in Listing~\ref{list1}, an analysis of the HTTP/2 protocol stack implementation in Chromium reveals that when two domains share the same Subject Alternative Name (SAN) \cite{rfc5280} and are resolved to the same IP address via DNS, the protocol stack may reuse a TCP connection. This mechanism changes the allocation of resources to different flows, affecting resource composition patterns within a flow. To simulate this phenomenon, we propose a SAN-constrained resource remapping method. The overall procedure is illustrated in Fig.~\ref{fig:overview}.
Specifically, we first extract SAN information from raw traffic and establish a mapping from resources to their registered domains and corresponding SAN sets. Next, we analyze the distribution of the number of IP nodes associated with each SAN in traffic traces and model it using a Gaussian distribution parameterized by the mean $\mu_{san}$ and standard deviation $\sigma_{san}$. During the data augmentation phase, given an input trace, we sample a target number of IP nodes $N$ from the distribution $\mathcal{N}(\mu_{san}, \sigma_{san})$, and reassign the $M$ domains within the same SAN set in the trace to these $N$ IP nodes. This process modifies the domain-to-IP mapping under protocol constraints, thereby adjusting how resources are distributed across concurrent flows and effectively augmenting the resource composition patterns within the original trace.

Furthermore, to address variations in flow reuse patterns for resources within the same domain, we design an empirically driven resampling method to augment flow reuse patterns. In the offline phase, we analyze historical traffic to collect allocation patterns of resources within the same domain across different flows, remove duplicate patterns, and construct a domain-specific pool of flow reuse patterns along with their empirical probability distributions. As illustrated in Fig.~\ref{fig:overview}, for a domain containing resources ${C, D}$, two reuse patterns may exist in historical observations: either ${C, D}$ are transmitted within the same flow, or ${C}$ and ${D}$ are distributed across two different flows.
In the online augmentation phase, given an input trace, we first extract the set of resources associated with the test set and search for matching historical reuse patterns with the same resource set in the related pattern pool. If matching patterns exist, one is randomly sampled according to its empirical probability and used to reorganize the current resource composition. If no matching pattern exists, a flow reuse pattern is randomly constructed for the current resource set under the constraint of a maximum number of concurrent flows.

In summary, these two resource recomposition strategies effectively simulate the variations in resource composition induced by the interplay between dynamic DNS and protocol stack reuse. By enriching the boundaries of the data distribution, they encourage the model to learn more generalizable representations of resource semantic grouping.


\subsection{Frame Sequence Augmentation}
\label{subsec:frame_augment}




We observe that the total upstream and downstream traffic volumes exhibit pronounced multimodal distributions, driven by factors such as \textit{HPACK} index hit states, accompanied by a clear static–dynamic separation in frame sequences, where some frame lengths remain stable while others vary within bounded ranges (Appendix~\ref{appendx_c2}).

Based on the above observations, we propose a frame sequence augmentation method with structure preservation and distributional constraints. It aims to generate HTTP frame sequences that conform to historical statistical distributions while retaining the stable structural patterns of application-layer frame sequences. Given the historical set of HTTP frame sequences for a certain resource, denoted as $\mathcal{S}$, each sequence can be represented as $S = [s_1, s_2, \dots, s_L]$, where positive values indicate upstream requests and negative values indicate downstream responses.

First, the method aligns the sample sequences in $\mathcal{S}$ and identifies positions where frame sizes remain nearly constant as anchor positions, which preserve the stable structural patterns of the resource’s frame sequence. Meanwhile, it detects the set of positions $\mathcal{M}$ where frame sizes exhibit variability, which are treated as adjustable positions for subsequent augmentation. For each adjustable position $i \in \mathcal{M}$, the historical variance $\sigma_i^2$ and value range $[b_i^{min}, b_i^{max}]$ are estimated to constrain the magnitude of local perturbations. In addition, for each historical sequence, the total upstream volume $U$ and downstream volume $D$ are computed, and their probability distributions $\hat{f}_U(u)$ and $\hat{f}_D(d)$ are estimated via kernel density estimation (KDE), capturing the global traffic volume variation patterns of the resource.

During the generation phase, the algorithm randomly selects a historical sequence $S^{base} \in \mathcal{S}$ as the base sequence, and samples a target upstream volume $U^{tgt}$ from the KDE distribution (the downstream sequence is generated analogously). The target volume is then allocated across the adjustable positions. Specifically, let the adjustable upstream frame vector to be generated be $\mathbf{x} = [x_1, \dots, x_k]^T$, with base values $\mathbf{x}^{base}$. This process is formulated as a constrained quadratic programming problem: under the constraints of total volume conservation and per-position value bounds, the objective is to minimize the deviation from the historical distribution weighted by variance. The optimization problem can be formally expressed as follows:

\begin{equation}
\begin{aligned}
\underset{\mathbf{x}}{\min} \quad 
& \sum_{i=1}^k \frac{(x_i - x_i^{base})^2}{\sigma_i^2 + \epsilon} \\
\text{s.t.} \quad 
& \sum_{i=1}^k x_i = U^{tgt}, \\
& b_i^{min} \le x_i \le b_i^{max}, \quad \forall i
\end{aligned}
\end{equation}

where $\epsilon$ is a small constant introduced to prevent division by zero. After solving the above optimization problem using the Sequential Least Squares Programming algorithm \cite{gill2011sequential}, we further incorporate a greedy heuristic to discretize the continuous solution $\mathbf{x}$ and correct residual errors. This procedure produces augmented sequences for individual resources that conform to historical statistical patterns in both local frame size variations and overall traffic volume.

After performing application-layer frame sequence augmentation, we further introduce a forward temporal shifting mechanism to simulate the combined transmission behavior of request \texttt{HEADER} frames under HTTP/2 multiplexing. Given a flow-level frame sequence $\mathcal{S}^{flow}$, formed by aggregating the frame sequences of all resources within a flow, we define $\mathcal{S}^{flow} = [S_1, S_2, \dots, S_N]$, where $S_t = [s_{t,1}, s_{t,2}, \dots, s_{t,L_t}]$ denotes the frame sequence of the $t$-th resource. The algorithm then processes these sequences in a backward manner.
For any request \texttt{HEADER} frames $s_{t,j} > 0$ in $S_t$ (for $t > 1$), the algorithm removes it from the current sequence with probability $p_{move} = 0.2$ and inserts it after the request \texttt{HEADER} frames in the preceding resource sequence $S_{t-1}$. Frames that have been shifted forward may continue to move further upstream in subsequent iterations with probability $p_{move}$, forming a cascading forward-shifting process.
Through this mechanism, requests \texttt{HEADER} from multiple resources can form localized aggregation patterns within the flow-level frame sequence, effectively simulating the coalesced transmission behavior induced by HTTP/2 multiplexing and mitigating the strict temporal patterns introduced by naive resource concatenation.

\subsection{Cross-Layer Feature Alignment}
\label{subsec:knowledge_distill}

\begin{figure}[h]
\centering
\includegraphics[width=\linewidth]{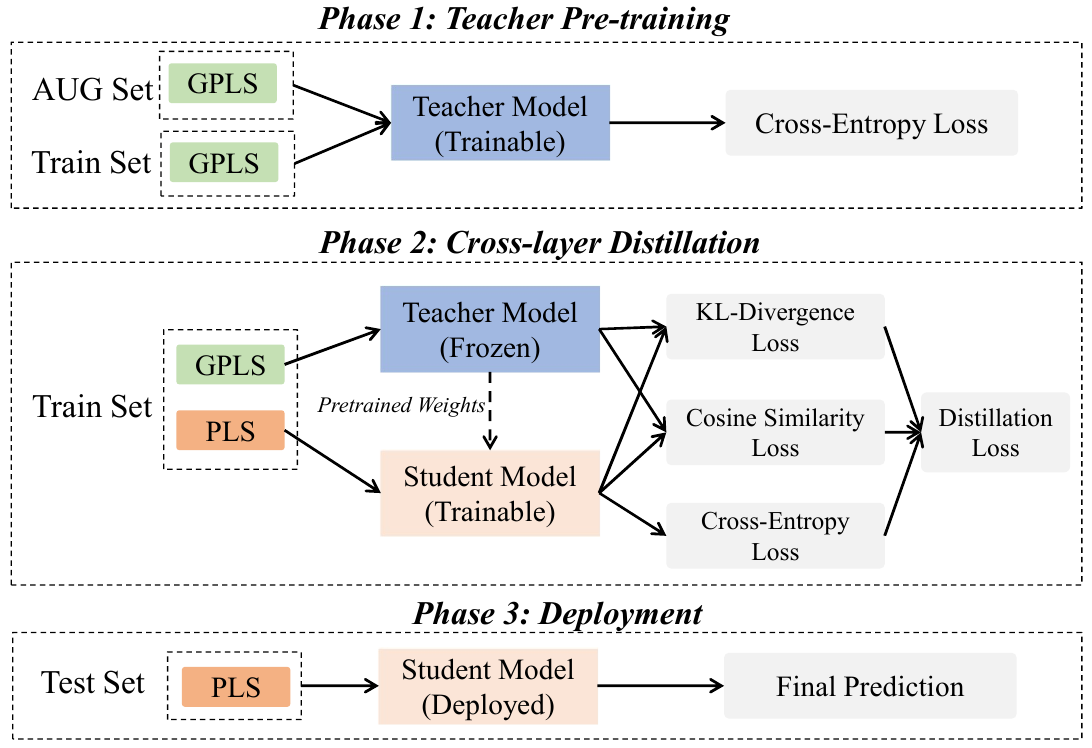}
\caption{Cross-layer feature alignment architecture.}
\label{fig:kd}
\end{figure}

Furthermore, to bridge the gap between application-layer frame sequences (FS) and transport-layer packet length sequences (PLS), we construct an intermediate proxy by generating an ideal packet length sequence from the frame sequence, and introduce a cross-layer feature alignment mechanism. In this mechanism, a teacher model equipped with frame-level semantic knowledge guides a student model to align its latent representations when processing observed packet length sequences. As a result, the model learns to suppress transmission perturbations and environmental noise.

First, we introduce an intermediate representation, termed the Generated Packet Length Sequence (GPLS), to mitigate the substantial discrepancies between FS and PLS in both feature dimensionality and value distribution. It represents a stable, idealized generation form of real traffic, free from transmission-induced perturbations such as buffering, fragmentation, and scheduling. In this form, each frame is independently encapsulated across protocol layers, yielding a packet-length structure that closely approximates an MSS-constrained segmentation pattern.
Let a single application-layer data frame be denoted as $f_i$, where $\text{sgn}(f_i)$ indicates the transmission direction and $|f_i|$ denotes its size. 
The reconstruction process consists of two sequential transformations. First, during cross-layer encapsulation, a fixed encapsulation and encryption overhead $\Delta_{TLS}$ is added to each application-layer frame to account for protocol stack overhead \footnote{$\Delta_{TLS}$ primarily consists of the fixed overhead introduced by the HTTP/2 frame header and the TLS Record layer. Under the widely used AES-GCM encryption mode, TLS 1.3 incurs approximately 31 bytes (including a 9B HTTP/2 frame header, 5B TLS Record header, 1B content type field, and 16B authentication tag), while TLS 1.2 introduces an additional $\sim$8B explicit initialization vector on top of this.}, yielding the ideal encapsulated frame $f'_i$:


\begin{equation}
\begin{aligned}
f'_i = \text{sgn}(f_i)(|f_i| + \Delta_{TLS})
\end{aligned}
\end{equation}

Subsequently, based on the MSS threshold $\tau_{MSS}$ of the target network environment, the encapsulated frame is subjected to idealized segmentation modeling. Specifically, when $|f'i| > \tau_{MSS}$, it is sequentially divided into a packet length subsequence $P_i$. In this process, $P_i$ consists of $k = \left\lfloor |f'i| / \tau_{MSS} \right\rfloor$ full-sized segments, along with an additional residual segment when the division is not exact, thereby characterizing the ideal segmentation structure.

\begin{equation}
\begin{aligned}
P_i = \text{sgn}(f_i) \big[ \underbrace{\tau_{MSS}, \dots, \tau_{MSS}}_{k}, |f'_i| \bmod \tau_{MSS} \big]
\end{aligned}
\end{equation}

As shown in Fig.~\ref{fig:kd}, based on the augmented data described above, we first perform supervised pretraining of a teacher model $\mathcal{T}$. It uses frame sequences to construct an ideal semantic representation space that is free from interference introduced by lower-layer transmission protocols. Given the GPLS $X_{\text{clean}}$ and corresponding labels $y$ from both the original training set and its augmented counterpart, the model parameters are optimized by minimizing a standard classification loss $\mathcal{L}_{cls}^{\mathcal{T}}$ (e.g., cross-entropy loss). After sufficient training, the teacher model can learn highly abstract and noise-robust semantic representations in its projection layer. Once training is complete, the parameters of the teacher model are frozen, and its output soft predictions and high-dimensional feature representations serve as reference targets for subsequent knowledge distillation \cite{gou2021knowledge}.

During the cross-layer distillation stage, the student model ($\mathcal{S}$) inherits the architecture and initialization of the teacher model and is trained on paired samples: the ideal sequence $X_{\text{clean}}$ (GPLS) is fed into the teacher model, while the corresponding noisy observed sequence $X_{\text{noisy}}$ (PLS) is fed into the student model. To guide the student model to approximate the semantic space under noisy inputs, we introduce a soft-label distillation mechanism in the decision space based on the Kullback–Leibler (KL) divergence. Let $z_t$ and $z_s$ denote the logits produced by the teacher and student models, respectively. A temperature parameter $T$ is applied to soften the output distributions, and the objective is to minimize the discrepancy between the two predictive distributions:

\begin{equation}
\mathcal{L}_{kl} = T^2 \cdot \mathcal{D}_{KL}\left( \sigma\left(\frac{z_t}{T}\right) \parallel \sigma\left(\frac{z_s}{T}\right) \right)
\end{equation}

where $\sigma(\cdot)$ denotes the Softmax function. This process enables the student model to capture the implicit inter-class structural relationships learned by the teacher model.
To further mitigate the impact of fragmentation and cross-layer perturbations on sequence representations, we introduce a cosine alignment constraint in the latent feature space. Specifically, let $v_t$ and $v_s$ denote the feature vectors output by the projection layers of the teacher and student models, respectively. Feature alignment is achieved by minimizing their cosine distance, thereby enforcing directional consistency between the PLS-based representation and the GPLS-based representation in the feature space.

\begin{equation}
\mathcal{L}_{cos} = 1 - \frac{1}{N} \sum_{j=1}^{N} \frac{v_s^{(j)} \cdot v_t^{(j)}}{\|v_s^{(j)}\|_2 \|v_t^{(j)}\|_2}
\end{equation}

Finally, by including the supervision from ground-truth labels, the overall optimization objective of the student model is defined as $\mathcal{L}_{student}$, where $\alpha$, $\beta$, and $\gamma$ are weighting coefficients. In this multi-objective joint optimization framework, the student model is able to progressively approximate the stable semantic representations characterized by GPLS, starting from perturbed PLS inputs. It enables more robust traffic fingerprinting performance in real-world network environments.

\begin{equation}
\mathcal{L}_{student} = \alpha \mathcal{L}_{cls}^{\mathcal{S}} + \beta \mathcal{L}_{kl} + \gamma \mathcal{L}_{cos}
\end{equation}

\section{Evaluation}

\subsection{Experimental Setup}

\paragraph{Dataset Construction}
To validate the effectiveness of {\sysname}, we construct an encrypted traffic dataset, as shown in Table~\ref{tab:dataset}, to evaluate model robustness in complex conditions such as cross-region and cross-time scenarios. Data collection is executed within Docker containers, where each URL visit is driven by an independent subprocess. Browser interactions are automated using Playwright \footnote{\url{https://playwright.dev/}}, while network traffic is captured by Tshark. In addition, TLS session keys are preserved to enable joint analysis of plaintext and encrypted traffic features.
The augmented dataset (AUG) is built by the Singapore-A dataset. The training set consists of 70\% of the Singapore-A data, with the remaining 15\% used for validation during training and 15\% reserved for testing. The training set contains both GPLS and their corresponding PLS, while the test set only includes observable PLS.
Importantly, to strictly evaluate the adaptability of {\sysname} under complex protocols such as HTTP/2, we use only HTTP/2 traffic in both training and evaluation stages.

\begin{table}[h]
\centering
\caption{Dataset Overview.}
\label{tab:dataset}
\resizebox{\linewidth}{!}{ 
\begin{tabular}{lccccc}
\toprule
\makecell[l]{Dataset \\ Name} & 
\makecell[c]{Collection \\ Date} & 
\makecell[c]{Collection \\ Location} & 
Description & 
\makecell[c]{Number of \\ Flows} & 
\makecell[c]{Number of \\ Traces} \\
\midrule
Singapore-A  & 2025/12 & Singapore   & Top-110 Alexa sites   & 160,604 & 5,500 \\
SouthKorea-A & 2025/12 & South Korea & Top-110 Alexa sites   & 175,472 & 5,500 \\
France-A     & 2025/12 & France      & Top-110 Alexa sites   & 70,703  & 5,500 \\
Singapore-B  & 2026/01 & Singapore   & Top-110 Alexa sites   & 156,753 & 5,500 \\
China-C      & 2025/03 & China       & Top-9853 China sites & 334,414 & 9,853 \\
\bottomrule
\end{tabular}}
\end{table}

\paragraph{Model and Hyperparameter Settings}
The experimental evaluation employs commonly used deep learning models for WF, including FSNet \cite{liu2019fs}, BERT-PS \cite{bertps}, Transformer, LSTM, and GRU. The main hyperparameters are configured as follows: the learning rate is set to $1\times10^{-4}$, and the maximum input sequence length is 500. Models are trained using the Adam optimizer. The maximum number of training epochs is set to 300, with early stopping terminating training if the validation F1 score does not improve for 15 consecutive epochs.

\paragraph{Experimental Environment}
To ensure reproducibility, all experiments are conducted on a consistent computing platform. The software environment comprises Ubuntu 22.04, Python 3.12, and PyTorch 2.7.0. The hardware configuration is anchored by an NVIDIA RTX 5090 GPU (32 GB) and a 32-core Intel(R) Xeon(R) Gold 6459C CPU. 

Our evaluation is structured to address the following research questions:

\begin{itemize}
\item \textbf{RQ1: Performance.}
Does {\sysname} significantly improve the classification accuracy and unknown-sample detection capabilities of representative baseline models across diverse scenarios?
\item \textbf{RQ2: Effectiveness of Key Components.}
How do the individual components of {\sysname} contribute to the overall performance enhancement?
\item \textbf{RQ3: Robustness and Generalization.}
Can {\sysname} maintain stable performance and resilience under highly volatile network conditions and complex deployment environments?
\item \textbf{RQ4: Efficacy under Controlled Settings.}
Does {\sysname} sustain its efficacy in controlled environments where dynamic noise is minimized?
\end{itemize}



\subsection{RQ1: Performance}

\begin{table*}[t]
\centering
\caption{Performance improvements of {\sysname} across different datasets and models.}
\label{tab:performance}
\resizebox{\linewidth}{!}{
\begin{tabular}{lcccccccc}
\toprule
\multirow{2}{*}{Methods} 
& \multicolumn{2}{c}{Singapore-A} 
& \multicolumn{2}{c}{France-A} 
& \multicolumn{2}{c}{SouthKorea-A} 
& \multicolumn{2}{c}{Singapore-B} \\
\cmidrule(lr){2-3} \cmidrule(lr){4-5} \cmidrule(lr){6-7} \cmidrule(lr){8-9}
& ACC & F1 & ACC & F1 & ACC & F1 & ACC & F1 \\
\midrule
Transformer 
& 73.24 (2.02\%$\uparrow$) & 84.45 (0.84\%$\uparrow$)
& 64.78 (2.81\%$\uparrow$) & 68.77 (2.44\%$\uparrow$)
& 52.53 (4.13\%$\uparrow$) & 68.73 (2.15\%$\uparrow$)
& 62.55 (1.74\%$\uparrow$) & 72.70 (1.28\%$\uparrow$) \\

LSTM 
& 65.49 (1.07\%$\uparrow$) & 79.93 (0.90\%$\uparrow$)
& 56.34 (3.39\%$\uparrow$) & 61.10 (2.49\%$\uparrow$)
& 44.69 (2.26\%$\uparrow$) & 58.54 (4.27\%$\uparrow$)
& 55.03 (1.07\%$\uparrow$) & 67.22 (2.01\%$\uparrow$) \\

GRU 
& 62.96 (0.83\%$\uparrow$) & 76.71 (0.87\%$\uparrow$)
& 53.91 (2.71\%$\uparrow$) & 59.43 (3.20\%$\uparrow$)
& 41.78 (3.78\%$\uparrow$) & 55.14 (6.86\%$\uparrow$)
& 52.71 (1.40\%$\uparrow$) & 65.42 (1.54\%$\uparrow$) \\

BERT-PS
& 83.92 (2.13\%$\uparrow$) & 93.38 (0.88\%$\uparrow$)
& 44.88 (10.98\%$\uparrow$) & 49.45 (7.14\%$\uparrow$)
& 39.91 (12.23\%$\uparrow$) & 50.43 (8.96\%$\uparrow$)
& 70.92 (2.55\%$\uparrow$) & 79.62 (1.82\%$\uparrow$) \\

FSNet 
& 71.07 (5.81\%$\uparrow$) & 82.77 (3.75\%$\uparrow$)
& 60.40 (5.76\%$\uparrow$) & 62.90 (9.01\%$\uparrow$)
& 47.81 (7.66\%$\uparrow$) & 61.79 (8.16\%$\uparrow$)
& 59.80 (5.43\%$\uparrow$) & 70.81 (4.28\%$\uparrow$) \\
\midrule
\textbf{On Average} 
& \textbf{71.34 (2.37\%$\uparrow$)} & \textbf{83.45 (1.45\%$\uparrow$)}
& \textbf{56.06 (5.13\%$\uparrow$)} & \textbf{60.33 (4.86\%$\uparrow$)}
& \textbf{45.34 (6.01\%$\uparrow$)} & \textbf{58.93 (6.08\%$\uparrow$)}
& \textbf{60.2 (2.44\%$\uparrow$)} & \textbf{71.15 (2.19\%$\uparrow$)} \\
\bottomrule
\end{tabular}
}
\end{table*}

\begin{figure*}[h]
\centering
\subfloat[\small ACC]{
    \includegraphics[width=0.31\linewidth]{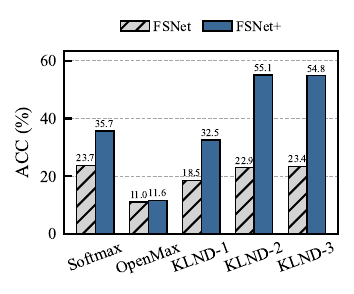}
    \label{subfig:ACC}
}
\hfill
\subfloat[\small F1-score]{
    \includegraphics[width=0.31\linewidth]{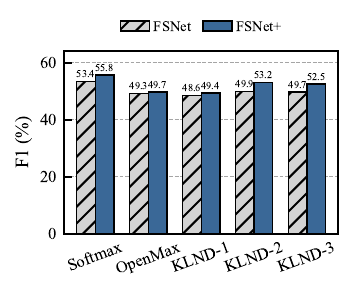}
    \label{subfig:F1}
    }
\hfill
\subfloat[\small AUROC]{
    \includegraphics[width=0.31\linewidth]{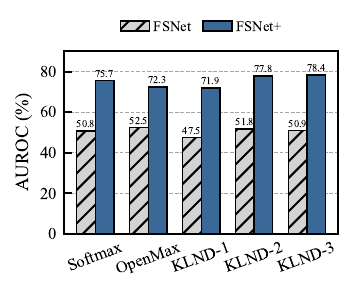}
    \label{subfig:AUROC}
    }
\caption{Performance improvements of {\sysname} across different open-world recognition mechanisms.}
\label{fig:openworld}
\end{figure*}

\paragraph{\textbf{Overall Performance}} Table~\ref{tab:performance} presents the performance comparison across different datasets and models. Overall, the proposed method consistently yields stable performance improvements across all evaluation scenarios and model configurations, validating its effectiveness in various settings.
In both cross-region scenarios (France-A and SouthKorea-A) and the cross-time scenario (Singapore-B), all models achieve varying degrees of performance improvement after incorporating the proposed method. In particular, under cross-region scenarios with more pronounced distribution shifts, the method achieves average improvements of 5.57\% and 5.47\% in ACC and F1, respectively, which are substantially higher than those observed in the closed-world setting (Singapore-A). It indicates that, by expanding resource composition patterns and mitigating structural perturbations, the proposed method effectively bridges the structural gap between source and target domains, thereby yielding more substantial gains under pronounced distribution shifts. From a model perspective, the method exhibits strong adaptability and architectural compatibility. Notably, larger improvements are observed in more expressive models such as BERT-PS and FSNet, with gains of up to 12.23\%. It suggests that more expressive models can better leverage the enriched semantic information in the augmented data when modeling packet-length sequences.

\paragraph{\textbf{Open-World Performance}} To evaluate the performance of {\sysname} in open-world scenarios, we use Singapore-A as the training set and conduct testing on France-A and China-C, where China-C is treated as unknown-class data to construct an open-world environment. During the inference stage, we further integrate several mainstream open-world recognition methods, including Softmax \cite{taylor2016appscanner}, OpenMax \cite{yang2021deep}, and a series of KLND-based methods \cite{dahanayaka2023robust} (KLND-1, KLND-2, and KLND-3). These methods are integrated with FSNet to systematically evaluate performance under different open-world decision mechanisms.

Fig.~\ref{fig:openworld} shows the performance improvements of FSNet in different open-world decision mechanisms after introducing the proposed method (FSNet+). The results show that, across all evaluation settings, the model achieves consistent improvements in ACC, F1, and AUROC, with particularly significant gains in ACC and AUROC, averaging 90.81\% and 48.37\%, respectively. It indicates that the proposed method not only enhances the model’s discriminative capability for known classes but also substantially improves its ability to identify unknown samples. In addition, the proposed method yields consistent performance gains across different open-world recognition methods, rather than relying on a specific decision strategy, further demonstrating its strong compatibility.

\begin{mybox}
\textbf{Answer to RQ1:} 
Experimental results demonstrate that {\sysname} consistently outperforms representative baselines across all scenarios. Notably, it achieves a 5.47\% average F1-score gain in cross-region deployments. In challenging open-world settings, {\sysname} yields remarkable improvements, boosting average Acc and AUROC by 90.81\% and 48.37\%, respectively.
\end{mybox}

\subsection{RQ2: Effectiveness of Key Components}

\begin{figure*}[t]
\centering
\subfloat[\small Singapore-A]{
    \includegraphics[width=0.31\linewidth]{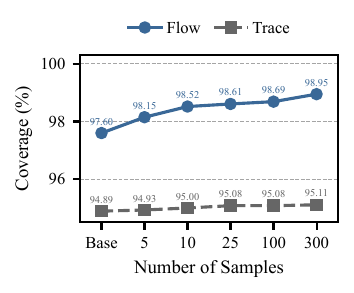}
    \label{subfig:singapore_coverage}
}
\hfill
\subfloat[\small SouthKorea-A]{
    \includegraphics[width=0.31\linewidth]{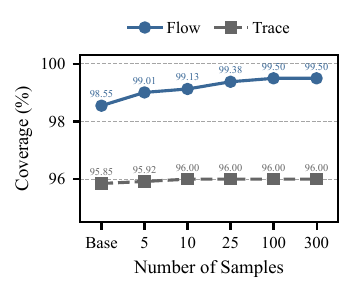}
    \label{subfig:suel_coverage}
    }
\hfill
\subfloat[\small France-A]{
    \includegraphics[width=0.31\linewidth]{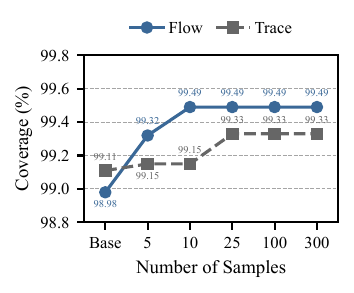}
    \label{subfig:france_coverage}
    }
\caption{Impact of resource recomposition on pattern coverage at flow and trace granularities.}
\label{fig:coverage}
\end{figure*}

\begin{figure*}[h]
    \centering
    \includegraphics[width=\linewidth]{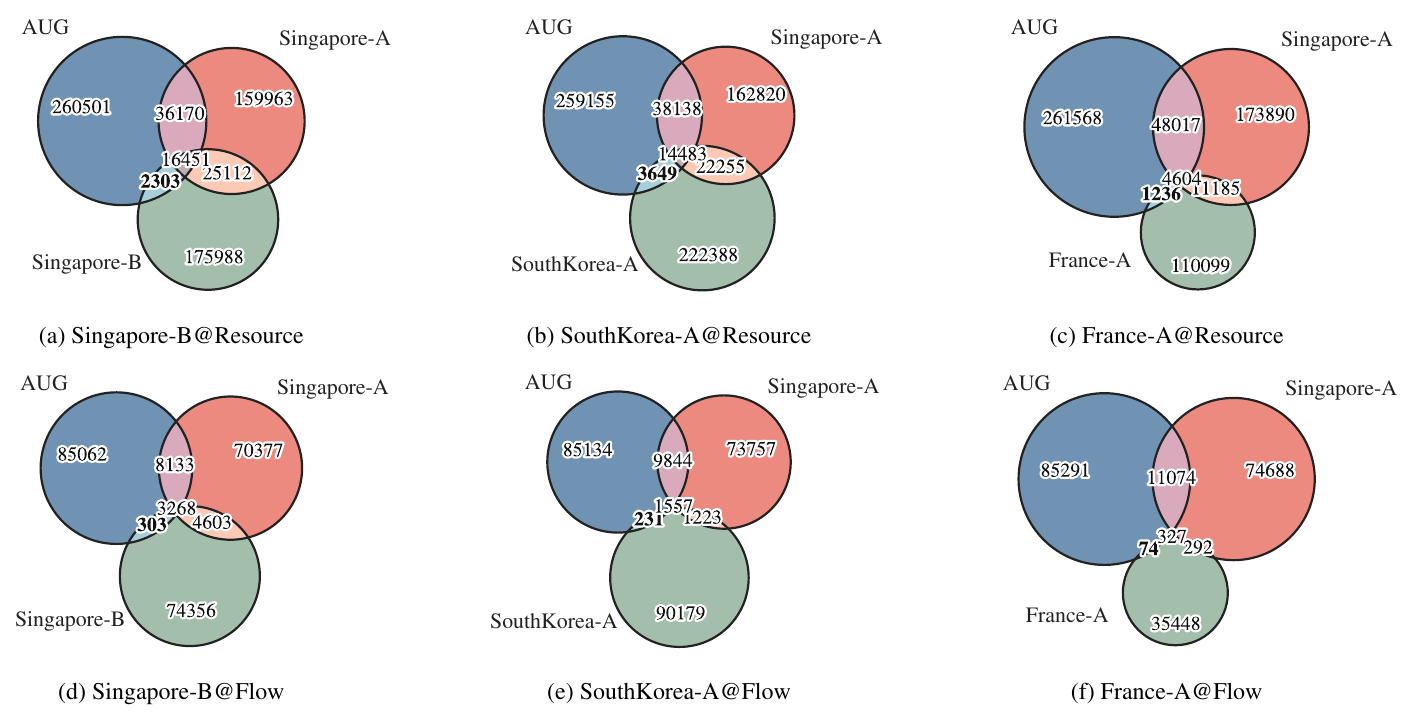}
    \caption{Impact of frame sequence augmentation on pattern coverage at resource and flow granularities.}
    \label{fig:Venn}
\end{figure*}

\begin{figure}[h]
    \centering
    \includegraphics[width=\linewidth]{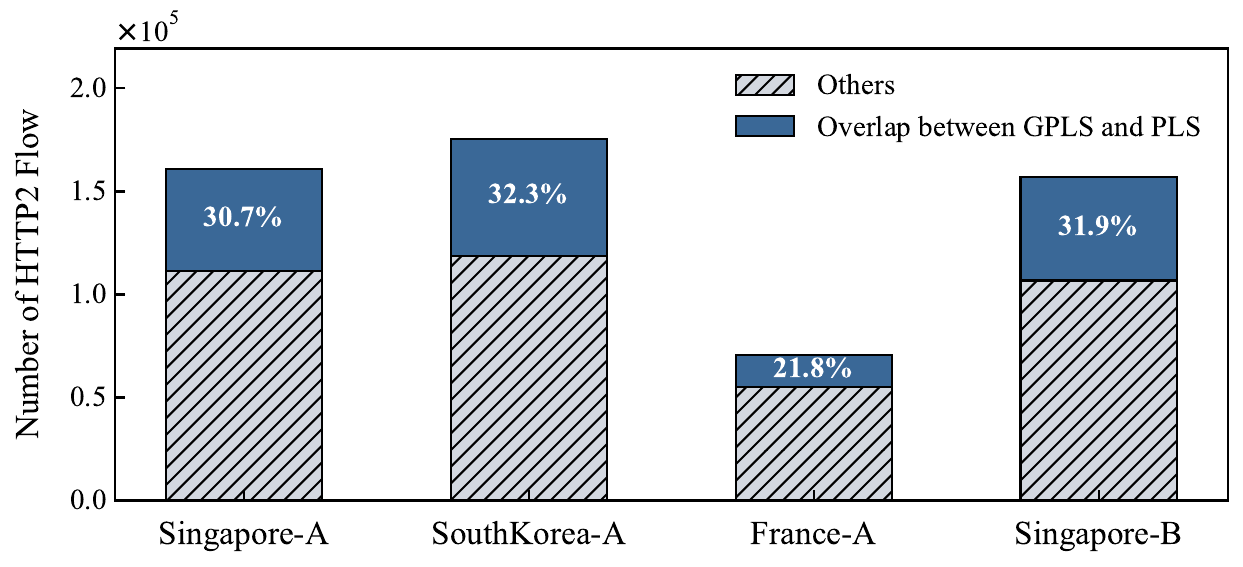}
    \caption{Overlap between GPLS and real PLS across datasets.}
    \label{fig:gpls_overlap}
\end{figure}


We evaluate the individual contribution of each key component to the overall performance through four distinct lenses: expansion of resource composition and frame sequence patterns, cross-layer feature alignment, and module ablation.

\paragraph{\textbf{Effect of resource recomposition}} To evaluate the effectiveness of the resource recomposition module, we remove the interference of dynamic resources and retain only a stable resource set. We then measure the coverage of resource composition patterns at both the flow and trace granularities. Specifically, we divide the dataset into train and test sets with an equal split, progressively increase the number of augmented samples generated, and compute the coverage between the resource composition patterns in the augmented training set and those in the test set.

The results in Fig.~\ref{fig:coverage} show that the coverage of resource composition patterns between the train and test sets increases as the number of augmented samples grows. On the Singapore-A, SouthKorea-A, and France-A datasets, the flow-level coverage reaches up to 98.95\%, 99.50\%, and 99.49\%, respectively, representing significant improvements over the baseline. In contrast, the improvement at the trace level is relatively limited. These results indicate that the proposed recomposition module can introduce richer composition patterns at the flow granularity, effectively expanding the pattern distribution of the training set.

It is worth noting that the above evaluation is based on a strict exact-match standard. Even in this constraint, the proposed method can generate resource composition patterns that are absent from the training set but genuinely present in the test set, demonstrating its ability to synthesize \textbf{`unobserved yet feasible'} structures. Furthermore, despite the relatively limited resource diversity in the current test set, the coverage still improves consistently, indicating that the method can effectively expand the coverage of composition patterns within a constrained distribution space. Given that resource compositions in real-world environments are typically more diverse and exhibit more complex variations, the proposed method is expected to provide greater advantages in highly diverse out-of-distribution scenarios.

\paragraph{\textbf{Effect of frame sequence augmentation}} We evaluate the effectiveness of the proposed frame-sequence augmentation method. Using Venn diagrams, we analyze the pattern coverage relationships among the augmented data, training set, and test set at two granularities: resource-level frame sequences, corresponding to individual resources, and flow-level frame sequences, corresponding to the aggregation of all resources within a flow. As shown in Fig.~\ref{fig:Venn}, the augmented dataset (AUG) constructed from Singapore-A is able to generate a large number of frame-sequence patterns that do not appear in the training set but genuinely exist in the test set, substantially expanding the data distribution.
Using the SouthKorea-A dataset as the test set, as shown in Fig.~\ref{fig:Venn}(b) and Fig.~\ref{fig:Venn}(e), the augmented dataset introduces an additional 3,649 resource-level frame sequence patterns and 231 flow-level frame sequence patterns, increasing the corresponding pattern coverage by 9.93\% and 8.31\%, respectively.  This result directly demonstrates that the proposed frame sequence augmentation method can effectively uncover and supplement latent structural variants, compensating for the insufficient coverage of training data in the sequence pattern space.

\paragraph{\textbf{Effect of feature alignment}} As shown in Fig.~\ref{fig:gpls_overlap}, we analyze the necessity and validity of the proposed generated packet length sequence. Results across the four datasets show that the generated packet length sequence (GPLS) from frame sequences exactly matches the real packet length sequence (PLS) in nearly 30\% of cases. This finding first indicates that the network protocol stack indeed introduces substantial nonlinear perturbations into PLS during transmission, rendering the mapping from application semantics to physical observations inherently uncertain. More importantly, it also validates the necessity and validity of the constructed GPLS, which, to some extent, captures the packet length generation process in the absence of transmission perturbations. Thus, it provides an effective approximation for bridging application semantics and physical observations.

\begin{figure}[h]
    \centering
    \includegraphics[width=\linewidth]{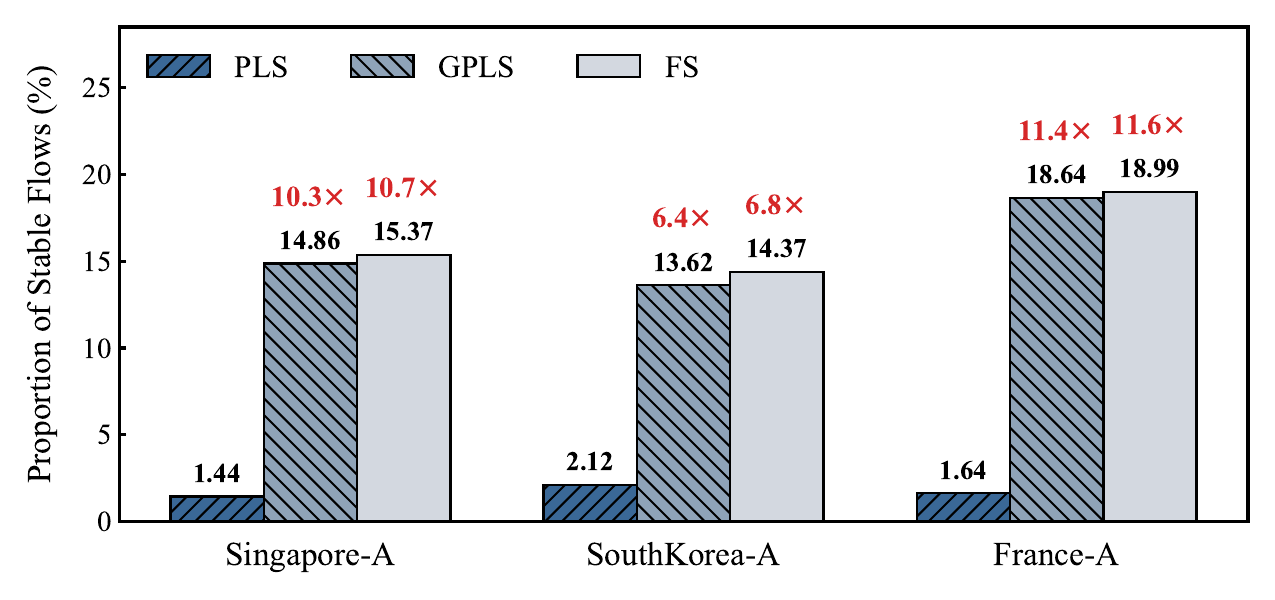}
    \caption{Robustness evaluation of different feature levels against transmission perturbations.}
    \label{fig:feature_stable_ratio}
\end{figure}

Fig.~\ref{fig:feature_stable_ratio} measures the stability of different feature representations while keeping the resource composition within each flow unchanged. The results show that, for PLS, only about 1.73\% of flows remain stable on average, indicating that PLS is highly sensitive to low-level transmission perturbations.  In contrast, higher-level feature representations, namely frame sequences (FS) and the GPLS, exhibit average stability levels approximately 9.37$\times$ and 9.06$\times$ that of PLS, respectively, demonstrating markedly stronger robustness. This finding confirms that higher-level representations are substantially more resilient to cross-layer perturbations and helps explain why low-level features alone are poor for stable representation learning. It further motivates our feature alignment method, which leverages application-layer semantic guidance to encourage more consistent and noise-robust PLS representations.

\begin{table}[h]
\centering
\caption{Performance comparison of different feature representations and ablation study.}
\label{tab:features}
\resizebox{\columnwidth}{!}{
\begin{tabular}{lcccc}
\toprule
\multirow{2}{*}{Feature/Methods} 
& \multicolumn{2}{c}{France-A} 
& \multicolumn{2}{c}{SouthKorea-A} \\
\cmidrule(lr){2-3} \cmidrule(lr){4-5}
& ACC & F1 & ACC & F1 \\
\midrule

GPLS+ 
& 64.93 (7.50\%$\uparrow$) & 71.28 (13.32\%$\uparrow$)
& 54.05 (13.05\%$\uparrow$) & 73.19 (18.45\%$\uparrow$) \\

FS 
& 65.78 (8.91\%$\uparrow$) & 70.49 (12.07\%$\uparrow$)
& 53.84 (12.61\%$\uparrow$) & 71.42 (15.59\%$\uparrow$) \\

GPLS 
& 62.86 (4.07\%$\uparrow$) & 69.23 (10.06\%$\uparrow$)
& 51.70 (8.14\%$\uparrow$) & 70.95 (14.82\%$\uparrow$) \\
\midrule

{\sysname} 
& 63.88 (5.76\%$\uparrow$) & 68.57 (9.01\%$\uparrow$)
& 51.47 (7.66\%$\uparrow$) & 66.83 (8.16\%$\uparrow$) \\

{\sysname} w/o RR 
& 64.18 (6.26\%$\uparrow$) & 67.73 (7.68\%$\uparrow$)
& 51.31 (7.32\%$\uparrow$) & 65.77 (6.44\%$\uparrow$) \\

{\sysname} w/o FSA 
& 62.66 (3.74\%$\uparrow$) & 66.4 (5.56\%$\uparrow$)
& 50.07 (4.73\%$\uparrow$) & 64.58 (4.52\%$\uparrow$) \\

{\sysname} w/o AUG 
& 62.79 (3.96\%$\uparrow$) & 66.5 (5.72\%$\uparrow$)
& 50.52 (5.67\%$\uparrow$) & 65.34 (5.75\%$\uparrow$) \\

\midrule
\rowcolor{gray!10}PLS 
& 60.40 & 62.90 & 47.81 & 61.79 \\

\bottomrule
\end{tabular}
}
\end{table}

Table~\ref{tab:features} presents the performance comparison of different features on FSNet. Compared with the PLS, both GPLS and FS yield significant performance gains across datasets, with FS achieving the largest F1 improvement of 15.59\% on SouthKorea-A. Overall, as the feature moves from PLS to GPLS and further to FS, model performance improves consistently, which aligns with the observation in Fig.~\ref{fig:feature_stable_ratio} that higher-level features exhibit stronger stability.
Furthermore, after joint training with GPLS generated from the AUG dataset, GPLS+ improves F1 by 13.32\% and 18.45\% over PLS on France-A and SouthKorea-A, respectively. It indicates that frame sequence augmentation can effectively expand the application-layer feature distribution and enhance its discriminative capability. Finally, although {\sysname} uses only PLS as input during testing, it still achieves an average F1 improvement of 8.56\% over PLS and approaches the performance of GPLS. It demonstrates that cross-layer feature alignment can guide observed features to approximate more stable application-layer semantic representations, improving overall classification performance.

\paragraph{\textbf{Ablation Study}} We conduct an ablation study to analyze the contribution of each key module to the overall performance. Specifically, we compare the full method ({\sysname}) with variants that remove the resource recomposition module (\texttt{w/o RR}), remove the frame sequence augmentation module (\texttt{w/o FSA}), and exclude augmented data (\texttt{w/o AUG}). In the \texttt{w/o AUG} setting, the teacher model is trained solely on GPLS derived from the original training set.

The results in Table~\ref{tab:features} show that the proposed method outperforms all ablation variants on both datasets, indicating that each module provides effective gains. Removing RR leads to a slight performance drop, suggesting that resource composition modeling helps expand high-level semantic patterns. Removing FSA causes a more pronounced degradation, demonstrating that frame sequence augmentation plays a critical role in expanding the frame sequence pattern space. Notably, \texttt{w/o AUG} still outperforms \texttt{w/o FSA}, indicating that, without FSA, naively assembled GPLS deviates substantially from the real distribution. It further verifies the importance of FSA in structural correction and realism enhancement.
Meanwhile, \texttt{w/o AUG} still significantly outperforms the PLS, showing that even without augmented data, the cross-layer feature alignment mechanism can effectively transfer high-level semantic knowledge to low-level features. Overall, RR, FSA, and the cross-layer feature alignment module improve representation stability and discriminative capability from the perspectives of pattern expansion and feature alignment, respectively, validating the effectiveness and rationality of the proposed design.

\begin{table*}[t]
\centering
\caption{Performance improvements of {\sysname} in few-shot settings.}
\label{fewshot}
\resizebox{\linewidth}{!}{
\begin{tabular}{lcccccccc}
\toprule
\multirow{2}{*}{Methods} 
& \multicolumn{2}{c}{Singapore-A} 
& \multicolumn{2}{c}{France-A} 
& \multicolumn{2}{c}{SouthKorea-A} 
& \multicolumn{2}{c}{Singapore-B} \\
\cmidrule(lr){2-3} \cmidrule(lr){4-5} \cmidrule(lr){6-7} \cmidrule(lr){8-9}
& ACC & F1 & ACC & F1 & ACC & F1 & ACC & F1 \\
\midrule

FSNet-3 
& 32.14 (30.96\%$\uparrow$) & 42.38 (25.79\%$\uparrow$)
& 27.84 (24.60\%$\uparrow$) & 32.79 (23.30\%$\uparrow$)
& 21.02 (29.88\%$\uparrow$) & 30.00 (28.87\%$\uparrow$)
& 26.48 (27.79\%$\uparrow$) & 36.45 (22.36\%$\uparrow$) \\

FSNet-10 
& 54.27 (17.91\%$\uparrow$) & 67.24 (13.37\%$\uparrow$)
& 45.12 (13.30\%$\uparrow$) & 49.83 (14.89\%$\uparrow$)
& 34.08 (18.87\%$\uparrow$) & 47.21 (19.76\%$\uparrow$)
& 43.35 (15.52\%$\uparrow$) & 55.39 (13.59\%$\uparrow$) \\

FSNet-20 
& 69.05 (1.04\%$\uparrow$) & 80.48 (1.14\%$\uparrow$)
& 53.95 (4.97\%$\uparrow$) & 58.72 (4.89\%$\uparrow$)
& 42.31 (6.88\%$\uparrow$) & 56.39 (8.19\%$\uparrow$)
& 52.87 (4.50\%$\uparrow$) & 65.62 (1.74\%$\uparrow$) \\

FSNet-50 
& 71.07 (5.81\%$\uparrow$) & 82.77 (3.75\%$\uparrow$)
& 60.40 (5.76\%$\uparrow$) & 62.90 (9.01\%$\uparrow$)
& 47.81 (7.66\%$\uparrow$) & 61.79 (8.16\%$\uparrow$)
& 59.80 (5.43\%$\uparrow$) & 70.81 (4.28\%$\uparrow$) \\

\midrule

\textbf{On Average} 
& \textbf{56.63 (13.93\%$\uparrow$)} & \textbf{68.22 (11.01\%$\uparrow$)}
& \textbf{46.83 (12.16\%$\uparrow$)} & \textbf{51.06 (13.02\%$\uparrow$)}
& \textbf{36.31 (15.82\%$\uparrow$)} & \textbf{48.85 (16.24\%$\uparrow$)}
& \textbf{45.63 (13.31\%$\uparrow$)} & \textbf{57.07 (10.49\%$\uparrow$)} \\

\bottomrule
\end{tabular}
}
\end{table*}

\begin{mybox}
\textbf{Answer to RQ2:} The results indicate that all three modules contribute substantially to performance gains. The first two reconstruct realistic yet unobserved resource composition and frame-sequence patterns, while the feature alignment module improves the stability of packet-length sequence representations.
\end{mybox}

\subsection{RQ3: Robustness and Generalization}
We further investigate the robustness and generalization of our method by analyzing its performance across several key dimensions: few-shot settings, feature granularity, class scale, and feature length.

\paragraph{\textbf{Few-Shot}} Table~\ref{fewshot} presents the data generation and feature enhancement capability of the proposed method in few-shot settings. The results show that, even in extremely data-scarce conditions, where only three access traces per website are used for training (FSNet-3), the proposed method remains effective and achieves significant performance gains. Across the four datasets, the average F1 improvement reaches 25.08\%, demonstrating the strong effectiveness of the proposed mechanism in clearing severe data scarcity. In addition, from the overall cross-dataset performance, the proposed method consistently yields substantial gains across all datasets, improving ACC by 12.16\%-15.82\% and F1 by 10.49\%-16.24\% on average. These results indicate that {\sysname} exhibits strong robustness and cross-environment generalization. Even when training samples are extremely limited, it can effectively enhance the model’s discriminative ability and representation quality, substantially reducing the model’s dependence on large-scale, high-quality labeled data.

\begin{table}[h]
\centering
\caption{Performance improvements across feature granularities.}
\label{udfs}
\resizebox{\linewidth}{!}{
\begin{tabular}{lcccc}
\toprule
\multirow{2}{*}{Methods} 
& \multicolumn{2}{c}{Singapore-A} 
& \multicolumn{2}{c}{France-A} \\
\cmidrule(lr){2-3} \cmidrule(lr){4-5}
& ACC & F1 & ACC & F1 \\
\midrule

UDFS         & 91.20 & 91.05 & 50.23 & 45.09 \\
PLS-Trace  & 90.82 & 90.74 & 27.42 & 23.66 \\
PLS   & 71.07 & 82.77 & 60.40 & 62.90 \\

\midrule

UDFS$^{+}$   & 91.56 & 91.33 & 54.49 & 47.52 \\
PLS-Trace$^{++}$  & 90.94 & 90.89 & 37.26 & 32.58 \\
PLS$^{++}$    & 75.20 & 85.87 & 63.88 & 68.57 \\

\midrule

\textbf{On Average}
& \textbf{84.36 (1.82\%$\uparrow$)} & \textbf{88.19 (1.33\%$\uparrow$)}
& \textbf{46.02 (12.73\%$\uparrow$)} & \textbf{43.88 (12.93\%$\uparrow$)} \\

\bottomrule
\end{tabular}
}
\end{table}


\paragraph{\textbf{Feature Granularity}} Table~\ref{udfs} presents the performance improvements of the proposed method at different granularities of features, including trace-level UDFS and PLS-Trace features, as well as flow-level PLS features. Among them, UDFS\cite{xian2025udfs} is a trace-level statistical sequence feature constructed by sequentially concatenating the upstream and downstream transmitted byte volumes of each flow within the same trace, while PLS-Trace globally merges all packet length sequences within the entire trace. It should be noted that PLS-based features are enhanced using the complete proposed method, denoted by $^{++}$, whereas UDFS is augmented only with the resource recomposition module, denoted by $^{+}$.

The results show that, after introducing {\sysname}, all types of features achieve consistent improvements in both ACC and F1 on the closed-world dataset (Singapore-A) and the cross-domain dataset (France-A), with the average improvement in the cross-domain scenario exceeding 12\%. This result fully demonstrates the effectiveness and generalizability of the proposed augmentation mechanism. Notably, since UDFS features are constructed solely through the recomposition of flow-level macroscopic statistics, the performance gains of UDFS$^{+}$ can be entirely attributed to the resource recomposition module. It further provides strong evidence that the module can generate plausible resource composition patterns and effectively expand the boundaries of the pattern space.

In addition, by comparing the performance of features at different granularities, we observe a notable phenomenon: trace-level features, including UDFS and PLS-Trace, perform well on the Singapore-A dataset, with F1 scores exceeding 90\%. However, they suffer from significant performance degradation on the cross-domain France-A dataset. For example, the F1 score of PLS-Trace drops to 23.66\%. A possible reason is that local distribution shifts at the flow level introduce error accumulation and sequence misalignment during global aggregation, causing substantial representation degradation and feature collapse for trace-level features in cross-domain scenarios.

\begin{table}[h]
\centering
\caption{Performance improvements in different numbers of classes.}
\label{class_num}
\resizebox{\columnwidth}{!}{
\begin{tabular}{lcccc}
\toprule
\multirow{2}{*}{Methods} 
& \multicolumn{2}{c}{Singapore-A} 
& \multicolumn{2}{c}{France-A}  \\
\cmidrule(lr){2-3} \cmidrule(lr){4-5}
& ACC & F1 & ACC & F1 \\
\midrule

FSNet@5 
& 95.45 (3.32\%$\uparrow$) & 95.51 (3.28\%$\uparrow$)
& 87.61 (3.86\%$\downarrow$) & 87.77 (0.60\%$\uparrow$) \\

FSNet@10 
& 90.54 (2.93\%$\uparrow$) & 91.42 (2.96\%$\uparrow$)
& 77.98 (2.33\%$\uparrow$) & 83.69 (1.05\%$\uparrow$) \\

FSNet@30 
& 81.94 (3.53\%$\uparrow$) & 85.99 (3.09\%$\uparrow$)
& 67.60 (1.79\%$\uparrow$) & 70.59 (4.87\%$\uparrow$) \\

FSNet@110 
& 71.07 (5.81\%$\uparrow$) & 82.77 (3.75\%$\uparrow$)
& 60.40 (5.76\%$\uparrow$) & 62.90 (9.01\%$\uparrow$) \\

\midrule

\textbf{On Average}
& \textbf{84.75 (3.90\%$\uparrow$)} & \textbf{88.92 (3.27\%$\uparrow$)}
& \textbf{73.40 (1.51\%$\uparrow$)} & \textbf{76.24 (3.89\%$\uparrow$)} \\

\bottomrule
\end{tabular}
}
\end{table}

\begin{table}[h]
\centering
\caption{Performance improvements in different sequence lengths.}
\label{feature_len}
\resizebox{\columnwidth}{!}{
\begin{tabular}{lcccc}
\toprule
\multirow{2}{*}{Methods} 
& \multicolumn{2}{c}{Singapore-A} 
& \multicolumn{2}{c}{France-A} \\
\cmidrule(lr){2-3} \cmidrule(lr){4-5}
& ACC & F1 & ACC & F1 \\
\midrule

FSNet-100 
& 65.60 (12.58\%$\uparrow$) & 77.12 (8.71\%$\uparrow$)
& 56.44 (8.13\%$\uparrow$) & 59.58 (8.68\%$\uparrow$) \\

FSNet-200 
& 66.71 (12.74\%$\uparrow$) & 79.58 (7.78\%$\uparrow$)
& 57.34 (11.58\%$\uparrow$) & 61.70 (10.47\%$\uparrow$) \\

FSNet-500 
& 71.07 (5.81\%$\uparrow$) & 82.77 (3.75\%$\uparrow$)
& 60.40 (5.76\%$\uparrow$) & 62.90 (9.01\%$\uparrow$) \\

\midrule

\textbf{On Average} 
& \textbf{67.79 (10.38\%$\uparrow$)} & \textbf{79.82 (6.75\%$\uparrow$)}
& \textbf{58.06 (8.49\%$\uparrow$)} & \textbf{61.39 (9.39\%$\uparrow$)} \\

\bottomrule
\end{tabular}
}
\end{table}

\paragraph{\textbf{Number of Classes}} Table~\ref{class_num} shows the impact of the number of classes $k$ on performance. As $k$ increases, stronger inter-class similarity and denser decision boundaries make the task more challenging, leading to performance degradation. Nevertheless, the proposed method consistently delivers stable gains across all scales, demonstrating robustness to class expansion. Notably, on the cross-domain France-A dataset, it achieves an average F1 improvement of 3.89\%, with larger gains at higher scales $k=110$. It indicates that the method effectively mitigates challenges such as data sparsity and feature overlap, exhibiting strong scalability in complex settings.

\paragraph{\textbf{Feature Length}} Table~\ref{feature_len} presents the performance under different PLS truncation lengths, where FSNet-100/200/500 denotes using the first 100, 200, and 500 packet lengths as input, respectively. Overall, as the truncation length increases, the model can access more complete sequence information, leading to gradual improvements in absolute performance. Meanwhile, the proposed method consistently brings stable gains at all length settings, achieving an average F1 improvement of 9.39\% on the France-A dataset. It indicates that the method can effectively improve model performance regardless of whether the input information is limited, demonstrating strong adaptability and robustness at different observation conditions and enabling stable effectiveness across multiple feature scales.

\begin{mybox}
    \textbf{Answer to RQ3:} 
    Experimental results demonstrate that {\sysname} consistently bolsters model performance under complex conditions, exhibiting superior robustness and generalization capabilities across volatile environments.
\end{mybox}

\subsection{RQ4: Efficacy under Controlled Settings}
We further construct a controlled dataset to eliminate the influence of dynamic application-layer resource variation and scheduling, enabling a more accurate evaluation of {\sysname}.

First, Table~\ref{stable_flow} ensures cross-domain resource homogeneity at the flow level, excluding noise introduced by resource composition variation with an ideal stable-flow setting. In this controlled condition, the overall performance of the baseline model improves substantially. 
Meanwhile, even in the generation setting without resource recomposition (\texttt{w/o RR}), the F1 improvements on the Singapore-A and SouthKorea-A datasets reach 2.1\% and 6.67\%, respectively, both of which are higher than those observed in the standard experiments in Table~\ref{tab:performance}. It indicates that, after excluding the influence of resource composition variation, the proposed method can effectively mitigate structural perturbations introduced during transmission by enhancing frame-sequence patterns and performing cross-layer feature alignment.

Furthermore, Table~\ref{stable_rescource} constructs an ideal stable-webpage setting in which the resource sets are strictly isomorphic across domains. In this clean environment, where interference from dynamic resources is excluded, the benefits of the proposed augmentation strategy are further increased, with the relative F1 gains on the two datasets increasing to 2.84\% and 8.54\%, respectively. This improvement beyond the results in Table~\ref{stable_flow} demonstrates the necessity of resource recomposition. It shows that, when the core resource pool of a webpage remains stable, the framework can efficiently generate plausible dynamic resource composition patterns, providing an effective solution for addressing the impact of resource composition variation.

\begin{table}[h]
\centering
\caption{Performance improvements in stable-flow setting.}
\label{stable_flow}
\resizebox{\columnwidth}{!}{
\begin{tabular}{lcccc}
\toprule
\multirow{2}{*}{Methods} 
& \multicolumn{2}{c}{Singapore-A} 
& \multicolumn{2}{c}{SouthKorea-A} \\
\cmidrule(lr){2-3} \cmidrule(lr){4-5}
& ACC & F1 & ACC & F1 \\
\midrule

Transformer 
& 76.09 (1.85\%$\uparrow$) & 84.53 (1.55\%$\uparrow$)
& 65.61 (5.75\%$\uparrow$) & 72.30 (4.92\%$\uparrow$) \\

LSTM 
& 67.31 (2.21\%$\uparrow$) & 78.66 (1.18\%$\uparrow$)
& 56.13 (4.49\%$\uparrow$) & 65.49 (3.65\%$\uparrow$) \\

GRU 
& 65.18 (1.46\%$\uparrow$) & 76.52 (1.36\%$\uparrow$)
& 53.64 (2.05\%$\uparrow$) & 59.82 (5.78\%$\uparrow$) \\

BERT-PS 
& 88.67 (1.29\%$\uparrow$) & 94.06 (0.85\%$\uparrow$)
& 41.24 (10.31\%$\uparrow$) & 45.98 (9.05\%$\uparrow$) \\

FSNet 
& 72.48 (8.06\%$\uparrow$) & 81.39 (5.57\%$\uparrow$)
& 60.48 (9.42\%$\uparrow$) & 65.31 (9.97\%$\uparrow$) \\

\midrule

\textbf{On Average}
& \textbf{73.95 (2.97\%$\uparrow$)} & \textbf{83.03 (2.10\%$\uparrow$)}
& \textbf{55.42 (6.40\%$\uparrow$)} & \textbf{61.78 (6.67\%$\uparrow$)} \\

\bottomrule
\end{tabular}
}
\end{table}

\begin{table}[h]
\centering
\caption{Performance improvements in stable-webpage setting.}
\label{stable_rescource}
\resizebox{\columnwidth}{!}{
\begin{tabular}{lcccc}
\toprule
\multirow{2}{*}{Methods} 
& \multicolumn{2}{c}{Singapore-A} 
& \multicolumn{2}{c}{SouthKorea-A} \\
\cmidrule(lr){2-3} \cmidrule(lr){4-5}
& ACC & F1 & ACC & F1 \\
\midrule

Transformer 
& 83.87 (1.48\%$\uparrow$) & 88.52 (0.98\%$\uparrow$)
& 74.60 (3.43\%$\uparrow$) & 77.31 (2.03\%$\uparrow$) \\

LSTM 
& 68.29 (6.57\%$\uparrow$) & 76.33 (5.03\%$\uparrow$)
& 58.13 (10.41\%$\uparrow$) & 61.15 (12.95\%$\uparrow$) \\

GRU 
& 72.34 (0.08\%$\uparrow$) & 78.96 (0.84\%$\uparrow$)
& 61.21 (2.19\%$\uparrow$) & 64.19 (3.79\%$\uparrow$) \\

BERT-PS 
& 93.50 (1.03\%$\uparrow$) & 96.14 (0.86\%$\uparrow$)
& 41.87 (15.64\%$\uparrow$) & 42.75 (13.82\%$\uparrow$) \\

FSNet 
& 78.23 (9.88\%$\uparrow$) & 83.90 (6.50\%$\uparrow$)
& 67.47 (12.41\%$\uparrow$) & 70.20 (10.10\%$\uparrow$) \\

\midrule

\textbf{On Average}
& \textbf{79.25 (3.81\%$\uparrow$)} & \textbf{84.77 (2.84\%$\uparrow$)}
& \textbf{60.66 (8.82\%$\uparrow$)} & \textbf{63.12 (8.54\%$\uparrow$)} \\

\bottomrule
\end{tabular}
}
\end{table}

\begin{mybox}
    \textbf{Answer to RQ4:} 
    The results under controlled settings demonstrate that the performance gains of {\sysname} originate from the intrinsic effectiveness of its module design, rather than being artifactually driven by environmental noise.
\end{mybox}

\section{Discussion and Future Work}
First, {\sysname} is specifically designed for mechanisms such as multiplexing and currently focuses on the HTTP/2 protocol. Given the prevalence of HTTP/2 in modern networks and our core focus on structural perturbations induced by multiplexing and cross-layer encapsulation, this specialization ensures a targeted and effective analysis in website fingerprinting.

Second, consistent with established literature like Rosetta \cite{xie2023rosetta}, {\sysname} primarily addresses the network transmission process. While it does not explicitly model the long-term evolution of web resources, the framework provides a robust foundation that can be extended to accommodate additional distribution shifts introduced by content changes.

Finally, while {\sysname} demonstrates exceptional efficacy on packet-length sequences, it currently concentrates on this primary side channel rather than auxiliary features like timing. Generalizing this cross-layer modeling to a multi-modal feature space represents a promising direction for future research.

\section{Conclusion}

This paper presents {\sysname}, a semantics-aware traffic augmentation framework for addressing the systematic shift between application semantics and observable traffic features in real-world environments. Under protocol constraints, {\sysname} expands resource-composition and frame-sequence patterns to enrich application-layer semantic representations, and introduces a cross-layer feature alignment module that distills enhanced semantics to align with observable packet-length features, improving representation stability.
Extensive evaluations demonstrate that {\sysname} not only synthesizes traffic patterns that are absent in the training set but present in the test set, but also consistently improves the performance of mainstream models across diverse and complex scenarios, validating its effectiveness in mitigating systematic feature shifts in realistic settings.




\appendices
\section{Stability Analysis of Resource Composition and Protocol}
\label{app:stability_analysis}

To further validate the objective existence of the discussed challenges and the rationale for focusing on HTTP/2-specific modeling, we provide additional statistical analysis on the stability of resource compositions and protocol-level behaviors.

\begin{figure}[h]
    \centering
    \includegraphics[width=\linewidth]{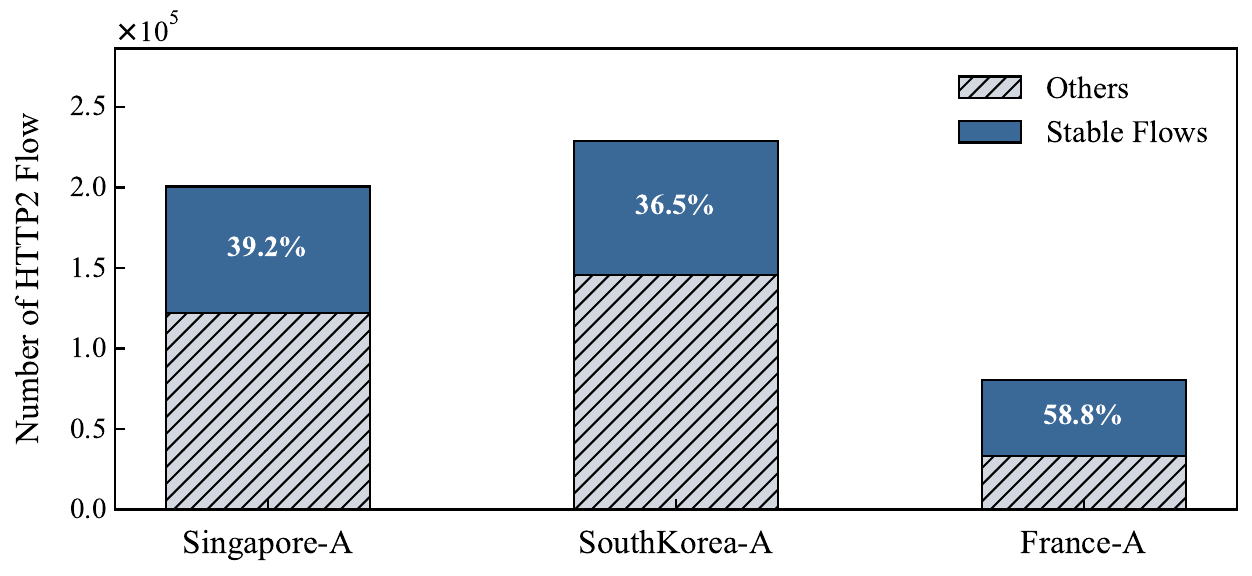}
    \caption{Stable flow ratio across datasets.}
    \label{fig:flow_stable_overlap}
\end{figure}

\begin{figure}[h]
    \centering
    \includegraphics[width=\linewidth]{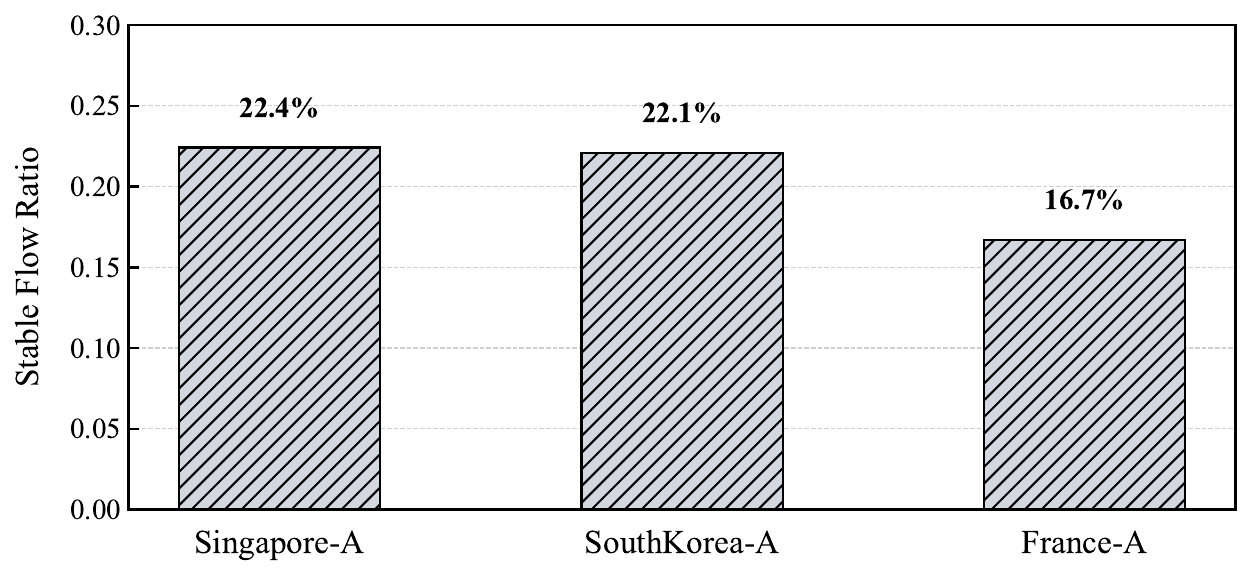}
    \caption{Resource consistency in stable subset.}
    \label{fig:flow_stable_in_stable_URI}
\end{figure}

\paragraph{Stability of resource composition within Flows}
We first examine the consistency of the resource set carried by a single flow for the same webpage across different traffic traces. As shown in Fig.~\ref{fig:flow_stable_overlap}, across all datasets, the proportion of stable flows, defined as flows whose resource composition remains similar across all traces, does not exceed 45\% on average. It indicates that even when accessing the same webpage, the set of resources within a flow can vary due to factors such as dynamic web content and network scheduling strategies.

To further eliminate the influence of intrinsic webpage content dynamics, we restrict the analysis to a subset of stable resources. As shown in Fig.~\ref{fig:flow_stable_in_stable_URI}, even when considering only these stable resources, more than 20\% of flows still exhibit inconsistent resource composition patterns on average. It provides strong empirical evidence for the existence of C1, demonstrating that the instability of in-flow resource compositions is not solely caused by content variation, but is also closely related to resource scheduling and protocol-level mechanisms.

\begin{figure}[h]
    \centering
    \includegraphics[width=\linewidth]{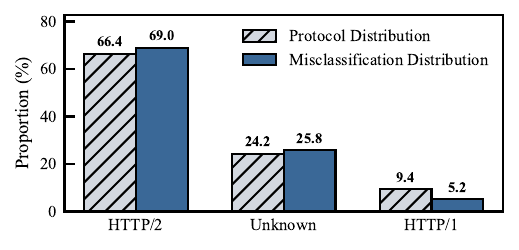}
    \caption{Protocol distribution in the dataset and misclassified samples.}
    \label{fig:protocol_distribution}
\end{figure}

\paragraph{Protocol-level Variability}
Finally, we analyze the distribution of different application-layer protocols and their impact on FSNet classification performance. As shown in Fig.~\ref{fig:protocol_distribution}, we present both the proportion of various TCP application-layer protocols across datasets and their distribution within misclassified samples. Notably, the Unknown category typically corresponds to pre-connection requests without an actual payload. The results show that HTTP/2 dominates overall traffic, which is consistent with existing measurement reports \footnote{\url{https://radar.cloudflare.com/en-us/year-in-review/2025}}. Meanwhile, HTTP/2 also accounts for a significantly higher proportion of misclassified samples compared to HTTP/1.

\begin{figure}[h]
    \centering
    \includegraphics[width=\linewidth]{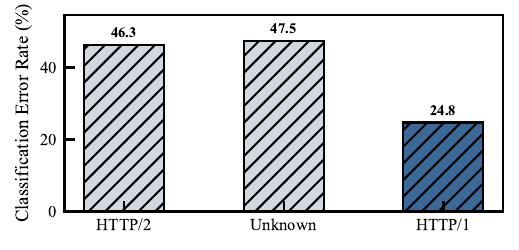}
    \caption{Classification error rate of different protocol.}
    \label{fig:classification_error_rate}
\end{figure}

Furthermore, as illustrated in Fig.~\ref{fig:classification_error_rate}, the classification error rate of HTTP/2 is substantially higher than that of HTTP/1, exceeding it by approximately 86.7\%. It indicates that HTTP/2 is not only the dominant protocol in the dataset but also the primary source of model misclassification. Its characteristics, such as multiplexing, dynamic header compression, and complex scheduling mechanisms, introduce stronger structural perturbations, making it more challenging for models to learn stable discriminative features. These findings highlight the necessity and rationality of focusing on HTTP/2-specific modeling and enhancement.

\section{resource composition Variability}
\label{appendx_c1}

\begin{figure}[h]
    \centering
    \includegraphics[width=\linewidth]{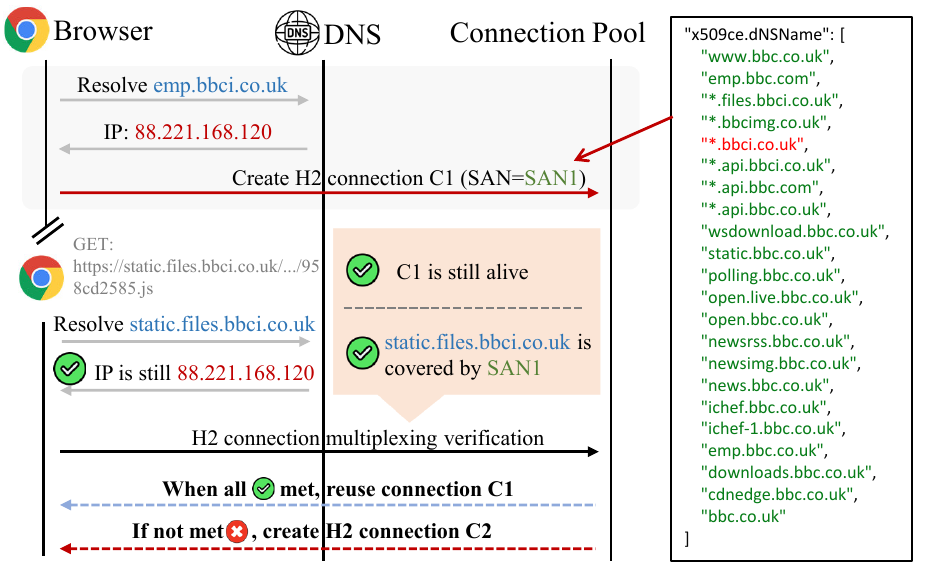}
    \caption{An illustration of the HTTP/2 connection coalescing mechanism. It demonstrates how cross-domain resources are aggregated into the same underlying TLS flow due to dynamic DNS resolution to the same IP and TLS certificate SAN matching.}
    \label{fig:case_cdn}
\end{figure}

To further intuitively illustrate the causes of resource composition variations within a flow, Fig. \ref{fig:case_cdn} provides an intuitive description of the interaction process between dynamic DNS and HTTP/2 connection coalescing.
In real-world networks, influenced by dynamic DNS mechanisms such as CDN scheduling, logically independent cross-domain requests are often resolved to the same physical edge server. For example, when a client establishes a TLS connection $C_1$ with \texttt{emp.bbci.co.uk}, the server typically provides a certificate containing a SAN list that covers multiple related domains. If the client subsequently requests resources from \texttt{static.files.bbci.co.uk} and the domain resolves to the same IP address, the browser performs HTTP/2 connection coalescing validation. If the connection $C_1$ remains active, the IP address matches exactly, and the new domain is included in the SAN list, the browser avoids establishing a new connection and instead reuses $C_1$ to transmit subsequent requests.

At the same time, even for resources within the same domain, requests are not always multiplexed over a single connection. This may occur when the number of concurrent streams reaches implementation limits, when the connection times out, or when either endpoint actively closes the connection, rendering the existing connection $C_1$ unavailable for reuse. As a result, subsequent requests from the same domain may be routed to new connections, leading to resource dispersion across multiple flows.

This protocol-level optimization, while designed to improve transmission efficiency, ultimately leads to two types of non-deterministic behaviors: flow-level aggregation of cross-domain resources and flow-level dispersion of same-domain resources. From the perspective of a traffic observer, such dynamic reorganization not only alters the actual resource set and traffic volume carried by a flow but also disrupts the stable mapping between application-layer webpage semantics and observation-layer packet length sequences. As a result, it introduces significant distribution shifts in the feature space. This mechanism fundamentally explains the origin of the application-layer resource composition variation generalization challenge discussed in this work.

\section{Frame Sequence Variations for a Single Resource}
\label{appendx_c2}


\begin{figure*}[h]
\centering
\subfloat[Resource A]{
    \includegraphics[width=0.45\linewidth]{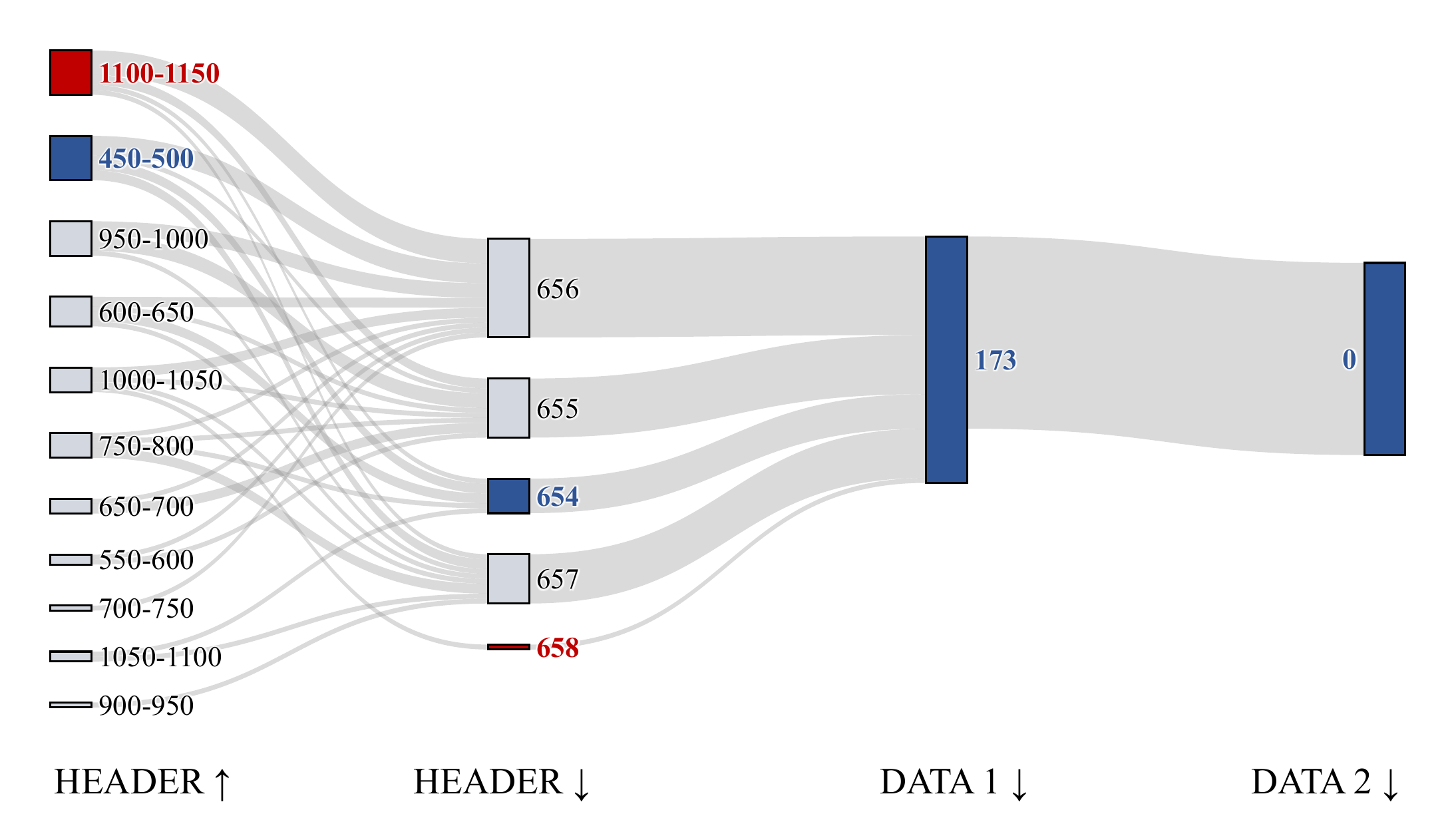}
    \label{fig:bfs1}
}
\hfill
\subfloat[Resource B]{
    \includegraphics[width=0.45\linewidth]{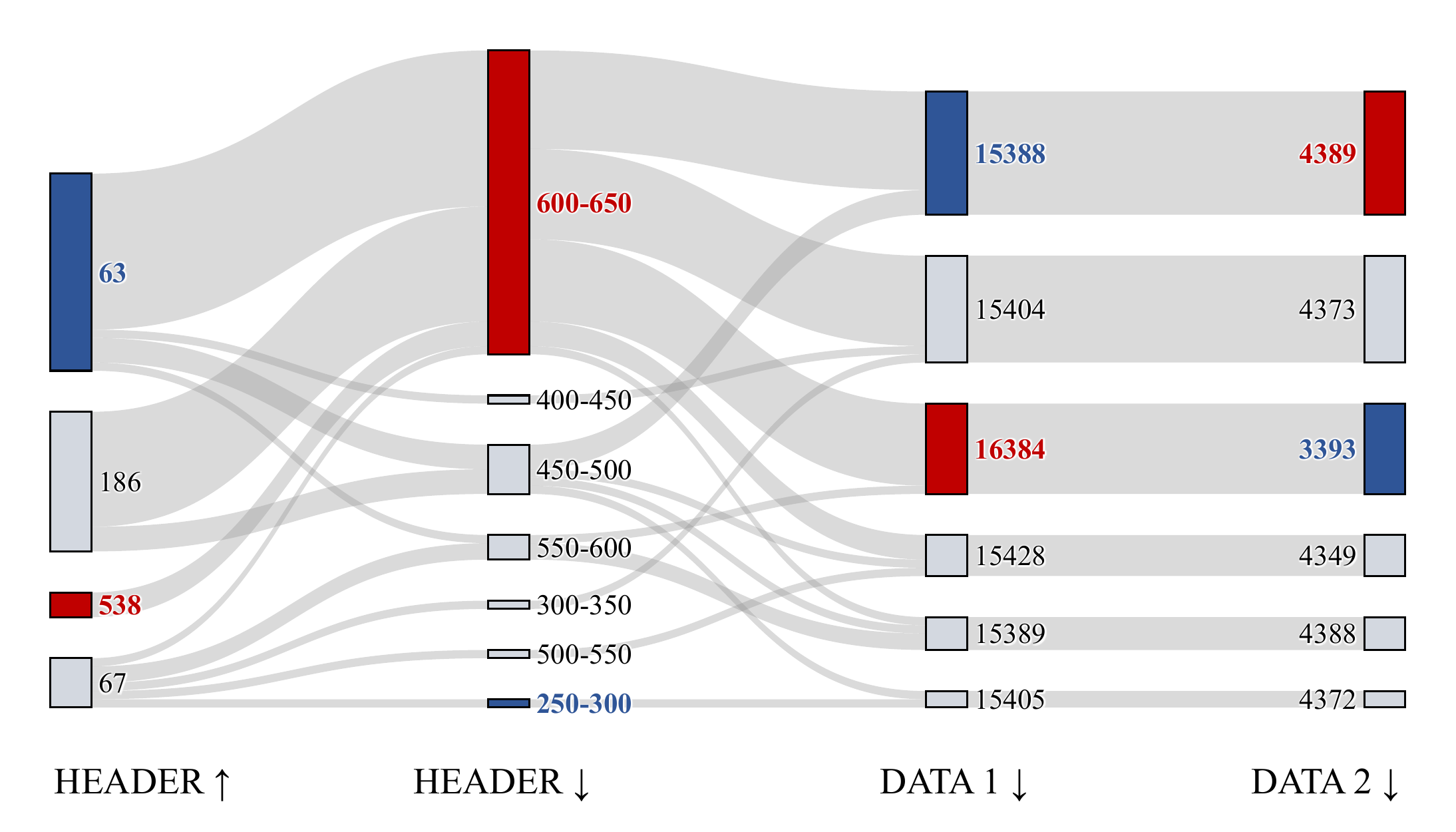}
    \label{fig:bfs2}
    }
\caption{Frame sequence variability for the same resource, showing stable structures and local perturbations across repeated accesses.}
\label{fig:bfs}
\end{figure*}

To further characterize the variability of frame sequences, we analyze them from both microscopic and macroscopic perspectives.

First, at the microscopic level, as shown in Fig.~\ref{fig:bfs}, the frame sequences corresponding to the same resource exhibit a clear static dynamic separation. On the one hand, some frames remain highly stable at specific temporal positions. For example, the DATA1 frame consistently maintains a fixed size of about 173 bytes across multiple observations, reflecting the stable encapsulation of fixed-structure data units at the protocol layer. On the other hand, frame lengths also exhibit nondeterministic variations. For instance, the server may append a 0-byte DATA frame at the end of transmission to indicate the end-of-stream state. In addition, during cross-layer encapsulation, the same logical payload may be segmented into different data chunks across visits due to asynchronous I/O buffer scheduling at the server side and dynamic changes in the HTTP/2 flow-control window. As a result, frame lengths may vary within a certain range at local temporal positions, as illustrated by DATA1 and DATA2 in Fig.~\ref{fig:bfs2}. This characteristic, where stable structure and random perturbation coexist, leads to pronounced instability in frame sequences at the microscopic level. It also constitutes a key source of structural noise introduced during cross-protocol-stack encapsulation and scheduling.

\begin{figure*}[h]
\centering
\subfloat[Indexed Header Field]{
    \includegraphics[width=0.45\linewidth]{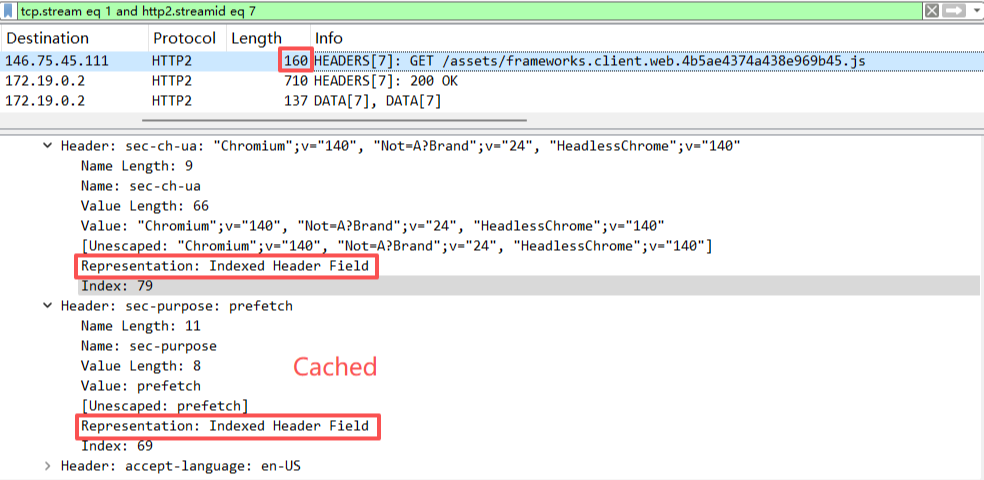}
    \label{fig:wireshark1}
}
\hfill
\subfloat[Literal Header Field]{
    \includegraphics[width=0.45\linewidth]{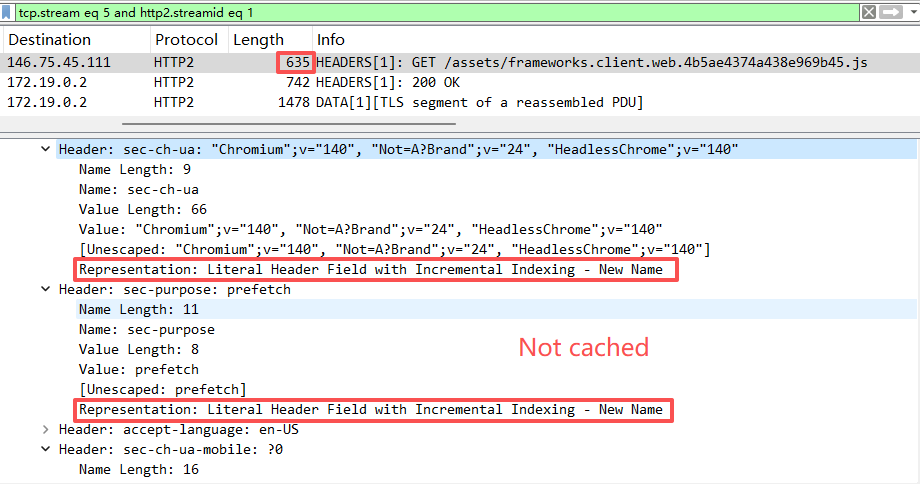}
    \label{fig:wireshark2}
    }
\caption{Wireshark-based analysis of HTTP/2 header compression, illustrating the impact of \textit{HPACK} states on frame size.}
\label{fig:wireshark}
\end{figure*}

Second, at the macroscopic level, as shown in Fig.~\ref{fig:bfs}, we observe that although frame sequences exhibit significant local changes at the microscopic level, their total upstream and downstream traffic volumes still approximately follow a multimodal distribution. Further analysis shows that this discretized distribution mainly arises from the state-dependent nature of \textit{HPACK} dynamic compression, together with variations in request parameters and header fields, Huffman encoding, and other factors.
The Wireshark \footnote{\url{https://www.wireshark.org/}} packet analysis of Fig.~\ref{fig:bfs2}, shown in Fig.~\ref{fig:wireshark}, further validates this phenomenon. When the \textit{HPACK} dynamic table cache is hit, as shown in Fig.~\ref{fig:wireshark1}, HTTP header fields can be efficiently compressed using \texttt{Indexed Header Field}, resulting in a \texttt{HEADERS} packet length of only 160 bytes. In contrast, during initial connection establishment or cache misses, as shown in Fig.~\ref{fig:wireshark2}, the headers must be explicitly transmitted using \texttt{Literal Header Field}, increasing the overhead of the same logical headers to 635 bytes.
This binary behavior caused by compression-state switching makes the transmission overhead of a single request jump among several discrete baseline values. When combined with minor perturbations introduced by dynamic fields and Huffman encoding, it ultimately leads to the multimodal distribution of total upstream and downstream traffic volumes at the macroscopic statistical level.

\section{Frame Sequence Variations for a Flow}
\label{appendx_c3}

\begin{figure}[h]
    \centering
    \includegraphics[width=\linewidth]{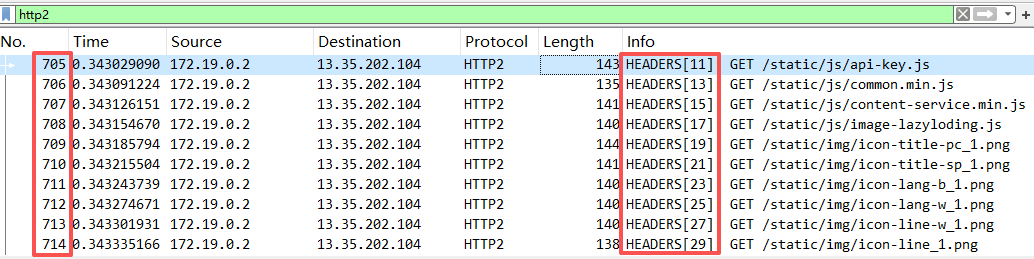}
    \caption{Concurrent transmission of continuous \texttt{HEADERS} frames driven by HTTP/2 multiplexing mechanism in Wireshark.}
    \label{fig:wireshark3}
\end{figure}

Fig.~\ref{fig:wireshark3} illustrates a commonly observed phenomenon in real HTTP/2 traffic, where HEADERS frames may appear in sub-millisecond bursts for certain resources. This is caused by HTTP/2 multiplexing together with constraints from the \textit{HPACK} header compression mechanism. When a client issues concurrent requests for multiple static resources, HTTP/2 assigns distinct stream IDs and transmits them over a shared TCP connection. To preserve \textit{HPACK} dynamic table consistency and comply with the requirement that each header block must be transmitted contiguously without interruption, the protocol stack schedules the corresponding HEADERS frames in a tightly packed and bursty manner.

%

\bibliographystyle{IEEEtran}
\bibliography{myref}

\end{document}